\documentclass{article}

\PassOptionsToPackage{numbers,sort&compress}{natbib}

\usepackage[final]{main}

\usepackage{wrapfig}
\usepackage[utf8]{inputenc} 
\usepackage[T1]{fontenc}    
\usepackage{hyperref}       
\usepackage{url}            
\usepackage{booktabs}       
\usepackage{amsfonts}       
\usepackage{nicefrac}       
\usepackage{microtype}      
\usepackage{xcolor}         
\usepackage{caption}
 \usepackage{silence}
\usepackage{amsmath}
\usepackage{amssymb}
\usepackage{mathtools}
\usepackage{amsthm}
\usepackage{multirow}
\usepackage[dvipsnames]{xcolor}
\usepackage{pifont}
\newcommand{\xmark}{\text{\ding{55}}}

\title{Bridging the Gap to Real-World Language-Grounded Visual Concept Learning}

\author{Whie Jung\hspace{2em}Semin Kim\hspace{2em}Junee Kim\hspace{2em}Seunghoon Hong \\
School of Computing, KAIST \\
\texttt{\{whieya, seminkim, kje0312, seunghoon.hong\}@kaist.ac.kr}
}

\begin{document}

\maketitle

\begin{abstract}
\vspace{-0.1in}
Human intelligence effortlessly interprets visual scenes along a rich spectrum of semantic dimensions. 
However, existing approaches to language-grounded visual concept learning are limited to a few predefined primitive axes, such as color and shape, and are typically explored in synthetic datasets.
In this work, we propose a scalable framework that adaptively identifies image-related concept axes and grounds visual concepts along these axes in real-world scenes. 
Leveraging a pretrained vision-language model and our universal prompting strategy, our framework identifies a diverse image-related axes without any prior knowledge.
Our universal concept encoder adaptively binds visual features to the discovered axes without introducing additional model parameters for each concept.
To ground visual concepts along the discovered axes, we optimize a compositional anchoring objective, which ensures that each axis can be independently manipulated without affecting others.
We demonstrate the effectiveness of our framework on subsets of ImageNet, CelebA-HQ, and AFHQ, showcasing superior editing capabilities across diverse real-world concepts that are too varied to be manually predefined. 
Our method also exhibits strong compositional generalization, outperforming existing visual concept learning and text-based editing methods. 
The code is available at \url{https://github.com/whieya/Language-grounded-VCL}.
\end{abstract}
\vspace{-0.18in}
\section{Introduction}
\vspace{-0.1in}
Perceiving the world through visual concepts such as color, shape, and texture, as human intelligence does, has long been a goal in computer vision.
Representing an image as a composition of these concepts not only improves compositional generalization~\cite{farhadi2009describe_by_attr,nagarajan2018attributes_as_operator,zhang-etal-2021-visually-grounded,zang2024vlm_discoverable_concept}, 
but also offers interpretable explanations~\cite{koh2020concept} and enhances visual reasoning tasks~\cite{gai-etal-2021-grounded-graph,zhong2024vis_table}.
Early work primarily used discrete language descriptors, ranging from object labels in classification and detection~\cite{russakovsky2015imagenet,krizhevsky2017imagenet,everingham2010pascal,lin2014mscoco} to sentence-level captions~\cite{vinyals2015show, anderson2018bottom}.
A recent method~\cite{lee2023livcl} shows that continuous concept embeddings, grounded along language-informed axes, can capture subtle visual nuances, \textit{e.g.}, slight color variations, beyond the reach of purely text-based approaches.
Thanks to the visual nuances embedded in continuous representations, 
this method enables the transfer of subtle, image-dependent details in downstream tasks such as image-editing tasks, where discrete text descriptor-based approaches~\cite{brooks2023instructpix2pix, mokady2023nulltext} often struggle due to limited linguistic expressiveness.

Despite this promise, extending the recent approach \cite{lee2023livcl} to learn diverse visual concepts in real-world scenes remains underexplored. 
A central challenge is the reliance on \emph{predefined} concept axes, 
such as color or shape, for visual grounding, which fails to capture the rich diversity of real images and limits extension to datasets where relevant factors are unknown in advance. 
Moreover, since each image consists of a wide variety of concept axes, relying on a specialized concept encoder for every axis quickly becomes infeasible, 
substantially increasing model complexity. 
Constraining each concept embedding to contain information relevant only to a specific concept axis presents another significant challenge. 
Although directly matching concept embeddings to textual descriptors—already a disentangled term in nature—offers a simple remedy for disentanglement \cite{lee2023livcl}, it compromises instance-specific details, as textual descriptors are image-agnostic.

In this work, we take a step toward a scalable approach for visual concept learning in real-world scenes. We leverage a pretrained vision-language model (VLM) to adaptively identify image-related axes, replacing fixed predefined ones. Using a universal prompt design, we guide the VLM to identify diverse image-related axes without relying on prior knowledge.
Our universal concept encoder then binds visual features to these discovered axes within a single unified architecture. To ensure that the discovered axes remain disentangled while preserving image-specific details, we introduce a compositional anchoring objective that constrains changes within each axis so that they only affect the corresponding axis in the generated images.
We demonstrate that our scalable framework can capture diverse real-world concepts and enable novel compositions of visual concepts.

In summary, our contributions are as follows:
\begin{enumerate}
\item We introduce a scalable framework that grounds visual concepts along diverse, language-specified axes in real-world images.
\item We propose adaptively identifying image-related axes with a pretrained VLM and designing a universal concept encoder that binds visual features to these axes.
\item We design a novel objective for disentangling discovered concept axes in real-world scenes.
\item We evaluate our framework on real-world concept editing tasks, showing superior editing capabilities and compositional generalization compared to language-informed visual concept learning methods and text-based editing methods.
\end{enumerate}
\vspace{-0.1in}

\section{Problem Setup}
\vspace{-0.08in}
Our goal is to develop a scalable framework for extracting visual concepts grounded along image-related linguistic axes in real-world images. 
To this end, we first outline a general formulation of language-grounded visual concept learning and identify the key challenges in scaling to real-world scenarios. Given an input image $\mathbf{x}\in\mathbb{R}^{H\times W\times C}$, 
the objective is to extract a set of concept representations $Z=\{\mathbf{z}_1,\dots,\mathbf{z}_K\}$, 
where $\mathbf{z}_i\in\mathbb{R}^D$ encodes visual concepts relevant to concept axis $y_i$.
To define interpretable axes among infinitely many concept axes in real-world images, we define each concept axis $y_i$ with natural languages, \textit{e.g.}, age, gender, and expression. 
Then the goal is to learn a set of concept encoders $E_{\theta_i}$ mapping $\mathbf{x}$ to visual concepts $\mathbf{z}_i$ corresponding to each concept axis $y_i$. 
A typical approach to train such encoders is training jointly with a decoder $D$ with an auto-encoding objective. 
The decoder $D$ is often replaced by a frozen pre-trained text-to-image (T2I) generative model \cite{lee2023livcl} due to training efficiency and remarkable generation capabilities. 
Formally, the encoders are optimized with the denoising objective: 
\begin{align}
\mathcal{L}_{\text{Diff}}(\{\theta_i\})=
\mathbb{E}_{\epsilon,t}\left[||D(\mathbf{x},t,\{E_{\theta_{i}}(\mathbf{x})\})-\epsilon||^2_2\right]
\label{eqn:recon}
\end{align}
where $\epsilon\sim\mathcal{N}(\textbf{0},I)$ and $t\sim U(0,1)$ denote noise and timestep, respectively. 

Since Equation~\ref{eqn:recon} does not guarantee the disentanglement of visual concepts along the concept axes, prior work~\cite{lee2023livcl} introduces additional regularization to ground each visual concept $\mathbf{z}_i$ to the text embeddings $\mathbf{v}_i$, which are obtained by querying the pretrained VLM ~\cite{li2023blip-2} with predefined templates, \textit{e.g.}, "what is the color of the object". 
We denote by $v_i$, \textit{e.g.}, red or blue, the textual descriptions for each axis, and define $\mathbf{v}_i=T(v_i)$ as their embeddings, where $T$ is a pretrained text encoder.

\subsection{Challenges and Desiderata}
\label{subsec:challenges and desiderata}
\vspace{-0.08in}
While prior work \citep{lee2023livcl} demonstrated the extraction of primitive visual concepts, \textit{e.g.}, color, shape, and style, primarily on simple synthetic datasets, extending this method to complex real-world scenes poses three key challenges.
We briefly outline these challenges and desiderata in this section, and discuss how they are addressed in Section~\ref{sec:approach}.

\paragraph{Adaptive Concept Axes}
\vspace{-0.08in}
Concept axes for visual grounding should be determined adaptively for each image, since real-world images exhibit a vast diversity of attributes that cannot be covered by a fixed set of predefined axes. 
Rather than relying on predefined primitive axes, \textit{i.e.}, color or shape, an adaptive mechanism is required to automatically identify relevant concept axes for each image.

\paragraph{Scalable Encoder Architecture}
To support adaptive concept axes, the encoder architecture should be scalable. Implementing $E_\theta$ with a set of specialized concept encoders for each concept axis $y_i$ would incur a prohibitive number of model parameters, considering infinitely many potential concept axes in real-world scenes. 

\vspace{-0.1in}
\paragraph{Concept Disentanglement}
Given adaptive concept axes, each representation $\mathbf{z}_i$ should capture only the semantics of its corresponding axis $y_i$, while preserving image-specific details.
A straightforward solution is to align $\mathbf{z}_i$ with the text embedding $\mathbf{v}_i$~\cite{lee2023livcl}, since texts are already disentangled along concept axes in nature. 
However, since $\mathbf{v}_i$ does not encode any instance-specific information, this alignment often leads to a suboptimal trade-off in $\mathbf{z}_i$ between encoding visual nuisance and disentanglement of the concepts. 

\vspace{-0.1in}
\section{Approach}
\label{sec:approach}
\begin{figure*}[t]
	\begin{center}
        \includegraphics[width=0.95\textwidth]{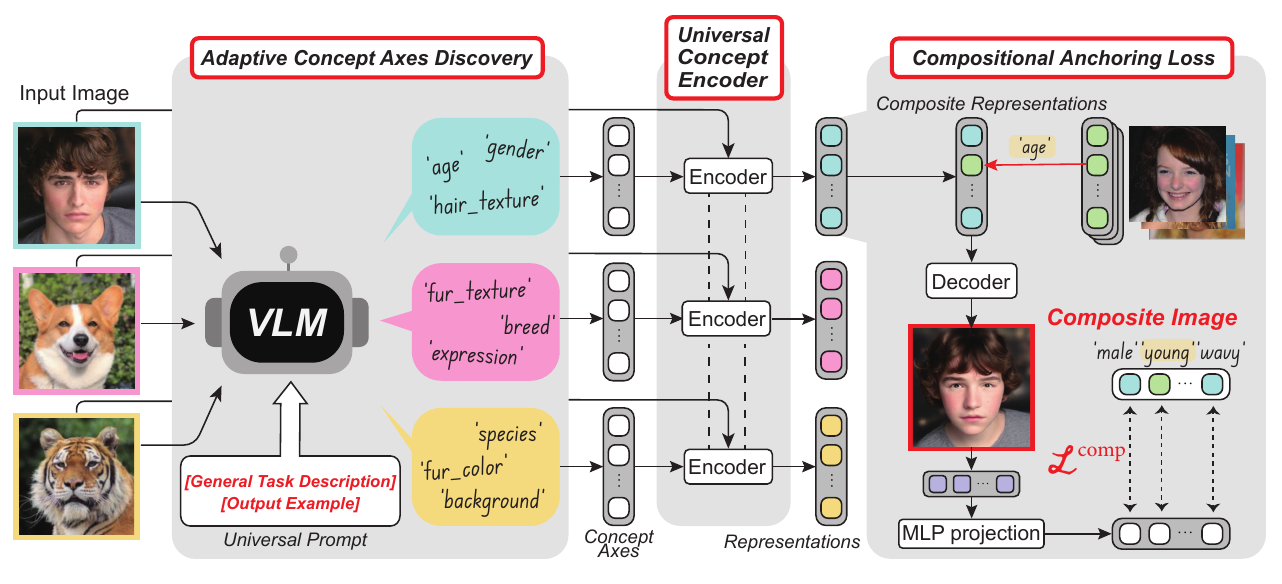}
        \vspace{-0.1in}
		\caption{
Overview of our method. Our framework first identifies image-related concept axes by leveraging VLM. We design an universal prompt that guides the VLM to find concept axes across different datasets. Given discovered axes for each image, our universal concept encoder binds visual features to those axes without introducing any specialized concept encoder for each axis. 
Finally, the encoded concept representations are regularized with a compositional anchoring loss to promote disentanglement between concept axes. Specifically, we randomly swap a concept representation with the one in an identical concept axis extracted from different images, and constrain composite images, rendered from randomly swapped representations, to be aligned with composite text descriptions. 
}
\label{fig:overview}
	\end{center}
\vspace{-0.25in}
\end{figure*}

\vspace{-0.1in}
Based on the desiderata outlined in Section~\ref{subsec:challenges and desiderata}, we present a scalable framework for language-grounded visual concept learning (Figure~\ref{fig:overview}). 
To extract concept axes adaptive to given images, we propose to leverage a pretrained VLM with our simple yet effective prompting strategy 
(Section~\ref{subsec:image-specific_axes}).
Given adaptive concept axes for each image, our universal concept encoder maps the image features to their corresponding visual concept embeddings
(Section~\ref{subsec:concept_encoder}).
We then train this encoder to disentangle visual concepts along the discovered axes by maximizing compositional anchoring of the representations (Section~\ref{subsec:compositional_anchoring}).
Instead of directly aligning  $\mathbf{z}_i$ with the image-agnostic text embedding $T(v_i)$, compositional anchoring ensures that changes in $\mathbf{z}_i$ affect only its corresponding concept axes $y_i$ in the generated image space $D(Z)$.
Below, we describe each component in detail.

\subsection{Adaptive Concept Axes Discovery}
\label{subsec:image-specific_axes}
\vspace{-0.05in}
Given an image $\mathbf{x}$, we query a pretrained VLM with a prompt $\mathcal{P}$ to extract image-dependent concept axes $Y=\{{y}_1, \dots, {y}_K\}$ and their corresponding textual descriptions $V=\{v_1,\dots, v_K\}$. 
Note that $K$ varies per image, and the extracted descriptions $V$ will be used for visual grounding in Section~\ref{subsec:compositional_anchoring}. 
The prompt $\mathcal{P}$ should be universal,  generalizing across arbitrary images and properly guiding the VLM to capture rich image-related concepts. 
To this end, we design a universal prompt with two key components: a \textbf{general task description} and an \textbf{output exemplar}.
The general task description instructs the VLM to enumerate all visually relevant concept axes presented in a given image. 
On the other hand, the \textbf{output exemplar} demonstrates the desired granularity of axes by providing a specific instance. 
By specifying axes in the exemplar, 
VLM can be steered to find more detailed axes, \textit{e.g.}, hair color, hair texture, and avoid overly coarse categories, \textit{e.g.,} color, texture. 
Remarkably, a single exemplar is sufficient to steer the VLM to identify 
diverse image-related concepts beyond those provided in the exemplar and to generalize to new domains. 
For example, given an instance of a human face that includes the axis 'hair color', the VLM discovers unspecified attributes such as 'eye color' or 'lip color' for different human faces, and identifies analogous axes for animal images, \textit{e.g.}, fur color. See Appendix~\ref{subsec:prompt_analysis} for more details.

We also instruct the VLM to structure the output as a dictionary mapping each concept axis to a corresponding textual description, \textit{e.g.} \{'age': 'young', 'gender': 'male',...\}. 
While prior work~\cite{lee2023livcl} employs the VLM to gather textual descriptions for a few predefined axes, 
they query the model separately for each predefined concept axis (e.g., “What is the color of the image?”). 
In contrast, our prompting strategy extracts all concept axes and corresponding textual descriptions in a single query, greatly enhancing efficiency and covering a broader range of potential axes beyond typical predefined categories. 
The complete prompt and outputs are provided in Appendix~\ref{subsec:prompt_analysis}. We find this universal prompt effectively captures diverse image-related concepts across multiple datasets, including novel concepts, \textit{e.g.}, breed, eye color, and nose color, which were \emph{not present} in the exemplar. 

\subsection{Universal Concept Encoder}
\label{subsec:concept_encoder}
\vspace{-0.1in}In our framework, the concept encoder $E_\theta$ requires to encode visual concepts adaptive to image-related concept axes $Y$.
Rather than defining specialized concept encoders for each concept, 
we construct $E_\theta$ to encode all concept representations $Z$ conditioned on a set of concept axes $Y$, \textit{i.e.}, $Z=E_\theta(\mathbf{x}, Y)$. 
The architecture for $E_\theta$ should support adaptive binding of visual features to given axes $Y$ and produce distinct concept representations within a single parameterized model.
To this end, we adapt the Querying Transformer (Q-Former)~\cite{li2023blip-2}, which was designed to extract visual features from a frozen vision encoder and align them with pretrained text embeddings. 
The Q-Former consists of a lightweight transformer with learnable queries to interact with visual features via a cross-attention module. 
In our adaptation, we replace the learnable queries with the text embeddings T($\{{y}_i\}$) of each axis encoded from a pretrained text encoder $T$. 
Initial queries $T(y_i)$ are then updated in subsequent transformer layers by interacting with visual features through cross attention layers. 
This way, visual features can dynamically bind to arbitrary concept axes within a single architecture. 

\subsection{Disentanglement with Compositionality}
\label{subsec:compositional_anchoring}
\vspace{-0.1in}
To constrain $\mathbf{z}_i$ to encode only the information relevant to its axis $y_i$, 
we introduce a \emph{compositional anchoring} objective that ensures modifying a concept along one axis alters the generated output only in that axis, leaving other attributes unchanged.
We implement such variations by randomly swapping a subset of concept representations 
$Z' \subseteq Z$ with those drawn from the same axis of different images, producing composite representations $Z^c$. 
As discovered axes vary across images, we first search for candidate images within each batch that share the same axis $y_i$, and then randomly swap their corresponding $\mathbf{z}_i$ among these candidates. 
When each representation $\mathbf{z}^i$ is disentangled along $y_i$, 
the composite image $\mathbf{x}^c=D(Z^c)$ should change only the swapped attributes, leaving others unchanged.
Since ground-truth images for such a composition are generally unavailable, we instead measure alignment between the composite image $\mathbf{x}^c$ and composed textual descriptions set $V^c$, constructed by taking the corresponding descriptions from each swapped axis. 

We quantify this alignment using a lightweight regression network $g_\phi$ that predicts the textual descriptions of a given image.
Instead of constructing $g_\phi$ with an additional image encoder,
we reuse $E_\theta$ to encode $\mathbf{x}^c$ back into concept representations, and a lightweight regression network $g_\phi$ predicts each attribute on top of the representations. Note that $g_\phi$ is shared across the axes. Formally, 
let ${\hat{Z}}^c = \{\mathbf{\hat{z}}^c_i\}_{i=1}^K=E_{\theta'}(\mathbf{x}^c,Y)$ be re-encoded concept representations from composite image $Z^c$, where $E_{\theta'}$ is a fixed copy of $E_\theta$. 
Then, the compositional anchoring objective is defined as: 
\begin{align}
\label{eqn:comp_anchor_loss}
\mathcal{L}_{\text{Comp}}(\theta)
= \sum_{i=1}^K d\Bigl(g_\phi\bigl(\hat{\mathbf{z}}^c_i\bigr),\, \mathbf{v}^c_i
\Bigr),
\end{align}
where $d(\cdot, \cdot)$ is a cosine distance and $\mathbf{v}^c_i$ is a text embedding for axis $y_i$ in $V^c$. 
Note that this objective only updates $\theta$ by propagating the gradient through $\mathbf{x}^c$ and prevents updating $g_\phi$ and $E_{\theta'}$ to avoid corruption from out-of-distribution samples of $D(\mathbf{z^c})$. 
For $g_\phi$, we simply train it by predicting the text embeddings $\mathbf{v}_i$ from non-swapped concept representations $\mathbf{z}_i$: 
\begin{align}
\label{eqn:regression_loss}
\mathcal{L}_{\text{Reg}}(\theta, \phi)
= \sum_{i=1}^K d\Bigl(g_\phi\bigl(\mathbf{z}_i),\, \mathbf{v}_i\Bigr),
\end{align}
It is worth noting that our objectives do not force $\mathbf{z}_i=\mathbf{v}_i$, which compromises instance-specific details in $\mathbf{z}_i$.
Instead, disentanglement is encouraged by verifying that each axis remains independent in the generated output. As a result, our objective ensures concept disentanglement while retaining instance-dependent information, particularly crucial in complex real-world scenarios.

\vspace{-0.05in}
\subsection{Learning objectives}
\vspace{-0.05in}
In this section, we summarize our complete framework and learning objectives. Given an image $\mathbf{x}$, the VLM extracts a set of image-related axes $Y$. 
The universal concept encoder $E_\theta$ is then trained with an autoencoding objective, while the pretrained decoder $D$ remains fixed. 
To encourage disentanglement among axes, we randomly swap each concept representation $\mathbf{z}_i$ with another from the same axis in the batch, and measure the alignment of composite image $\mathbf{x}^c = D(Z^c)$ with its corresponding text embeddings $\mathbf{v}^c$ through a lightweight regression network $g_\phi$.  The overall objective is:
\begin{align}
\mathcal{L}_{\text{Total}}(\theta, \phi)=
\mathcal{L}_{\text{Diff}}(\theta)+
\lambda_{\text{Comp}}\mathcal{L}_{\text{Comp}}(\theta)+
\lambda_{\text{Reg}}\mathcal{L}_{\text{Reg}}(\theta,\phi),
\label{eqn:total_obj}
\end{align}
where $\lambda_{\text{Comp}}$ and $\lambda_{\text{Reg}}$ are hyper-parameters controlling the importance of each term.

\vspace{-0.05in}
\section{Related Work}
\paragraph{Visual Concept Learning}
\vspace{-0.08in}
As language offers a human-interpretable interface, grounding visual concepts in natural language has long been a central goal in computer vision. 
Early efforts primarily aligned images with word-level annotations or object labels, supporting classification and detection tasks~\cite{krizhevsky2017imagenet,russakovsky2015imagenet,lin2014mscoco}. 
Extensions to neuro-symbolic frameworks \cite{mao2019neuro,li2020competence}, integrating with visual concept learning, further advanced visual reasoning. 
Such language-based grounding not only enhanced interpretability \cite{koh2020concept}, but also improved downstream performance on vision tasks \cite{li2022blip, li2023blip-2}. 
However, discrete text descriptors inherently limit the representational capacity to a fixed vocabulary. 
To address this, LIVCL \cite{lee2023livcl} followed Textual Inversion-based approaches~\cite{gal2022image} by optimizing concept encoders with a pretrained T2I model to reconstruct the given images. While promising, the scope of the work was limited to a few predefined primitive concept axes. 

\vspace{-0.1in}
\paragraph{Representation Learning with Compositionality}
Another line of research explores \emph{object-centric learning} to uncover generative factors. 
Recent methods \cite{wiedemer2023provable_ocl,jung2024l2c} compose latent representations from multiple images similar to our framework, but under more restrictive assumptions. 
For instance, L2C~\cite{jung2024l2c} randomly mixes object representations to produce composite images and maximizes the likelihood of these composites to learn object-centric representations. 
\citet{wiedemer2023provable_ocl} provides a theoretical analysis for compositional generalization and measures compositional consistency through a cyclic distance between latent representations and their reconstructions. 
However, this formulation relies on architectural constraints such as additive decoders, making them effective mainly on synthetic or low-complexity data. Without additive decoders, it can lead to a trivial solution where a single latent encodes all. 
In contrast, our approach employs a pretrained T2I model without imposing additional constraints, addressing real-world scenes with diverse concept axes. Instead of focusing on isolated objects, our compositional consistency objective promotes disentanglement among discovered concept axes and does not require a specialized decoder structure.

\vspace{-0.05in}
\section{Experiment}
\vspace{-0.08in}
\subsection{Experiment Setup}
\paragraph{Implementation Details}
\vspace{-0.08in}
We leverage InternVL~\cite{chen2024internvl} for an open-sourced VLM.
To handle complex real-world images, we employ DINO-v2~\cite{oquab2023dinov2} to encode the image into visual features followed by our concept encoder, and employ Stable Diffusion-based T2I decoder~\cite{rombach2022ldm} finetuned at 256$\times$256 resolution. 
When generating composite images with the T2I decoder, we iteratively decode for 10 steps using DDIM~\cite{song2020ddim}. 
Since propagating gradients through all these decoding steps is computationally expensive, we follow \cite{clark2023draft,prabhudesai2023aligning} and truncate gradients at the last few decoding iterations. Lastly, we employ 2-layer MLPs for $g_\phi$. See Appendix~\ref{subsec:additional_imp_detail} for additional implementation details.

\begin{figure*}[!t]
\centering
\begin{minipage}[!t]{\textwidth}
\tiny
\centering
\captionof{table}{
Comparisons on visual concept editing task. Our method consistently outperforms recent text-based editing methods~\cite{meng2021sdedit,mokady2023nulltext,brooks2023instructpix2pix,huberman2024ddpm_inversion} and language-informed visual concept learning~\cite{lee2023livcl}.  
} 
\vspace{-0.08in}
\label{table:overall_quat_values}
\begin{tabular}{@{}c|cc|cc|cc|cc@{}}
\toprule
                         & \multicolumn{2}{c|}{ImageNet-S20}                                             & \multicolumn{2}{c|}{CelebA-HQ}                                                & \multicolumn{2}{c|}{AFHQ-Dog}                                                 & \multicolumn{2}{c}{AFHQ-Cat}                                                  \\ \cmidrule(l){2-9} 
\multirow{-2}{*}{Method} & CLIP ($\uparrow$)               & BLIP ($\uparrow$)               & CLIP ($\uparrow$)               & BLIP ($\uparrow$)               & CLIP ($\uparrow$)               & BLIP ($\uparrow$)               & CLIP ($\uparrow$)               & BLIP ($\uparrow$)               \\ \midrule
SDEdit~\cite{meng2021sdedit}                   & {  0.195}          & {  0.381}          & {  0.200}          & {  0.447}          & {  0.250}          & {  0.493}          & {  0.257}          & {  0.474}          \\
InstructPix2Pix~\cite{brooks2023instructpix2pix}          & {  0.198}          & {  0.383}          & {  0.202}          & {  0.425}          & {  0.230}          & {  0.467}          & {  0.246}          & {  0.471}          \\
NullText Inversion~\cite{mokady2023nulltext}       & {  0.189}          & {  0.341}          & {  0.193}          & {  0.422}          & {  0.251}          & {  0.489}          & {  0.255}          & {  0.476}          \\
DDPM-Inversion~\cite{huberman2024ddpm_inversion}           & {  0.243}          & {  0.467}          & {  0.220}          & {  0.483}          & {  0.266}          & {  0.516}          & {  0.266}          & {  0.494}          \\
LIVCL~\cite{lee2023livcl}                    & {  -}              & {  -}              & {  0.226}          & {  0.469}          & {  0.270}          & {  0.518}          & {  0.268}          & {  0.480}          \\
Ours                     & {  \textbf{0.251}} & {  \textbf{0.474}} & {  \textbf{0.239}} & {  \textbf{0.496}} & {  \textbf{0.272}} & {  \textbf{0.535}} & {  \textbf{0.271}} & {  \textbf{0.514}} \\ \bottomrule
\end{tabular}

\end{minipage}

\begin{minipage}[!t]{\textwidth}
\scriptsize
\centering
\vspace{0.1in}
\begin{minipage}[!t]{1.0\textwidth}
\scriptsize
\centering
\includegraphics[width=0.9\linewidth]{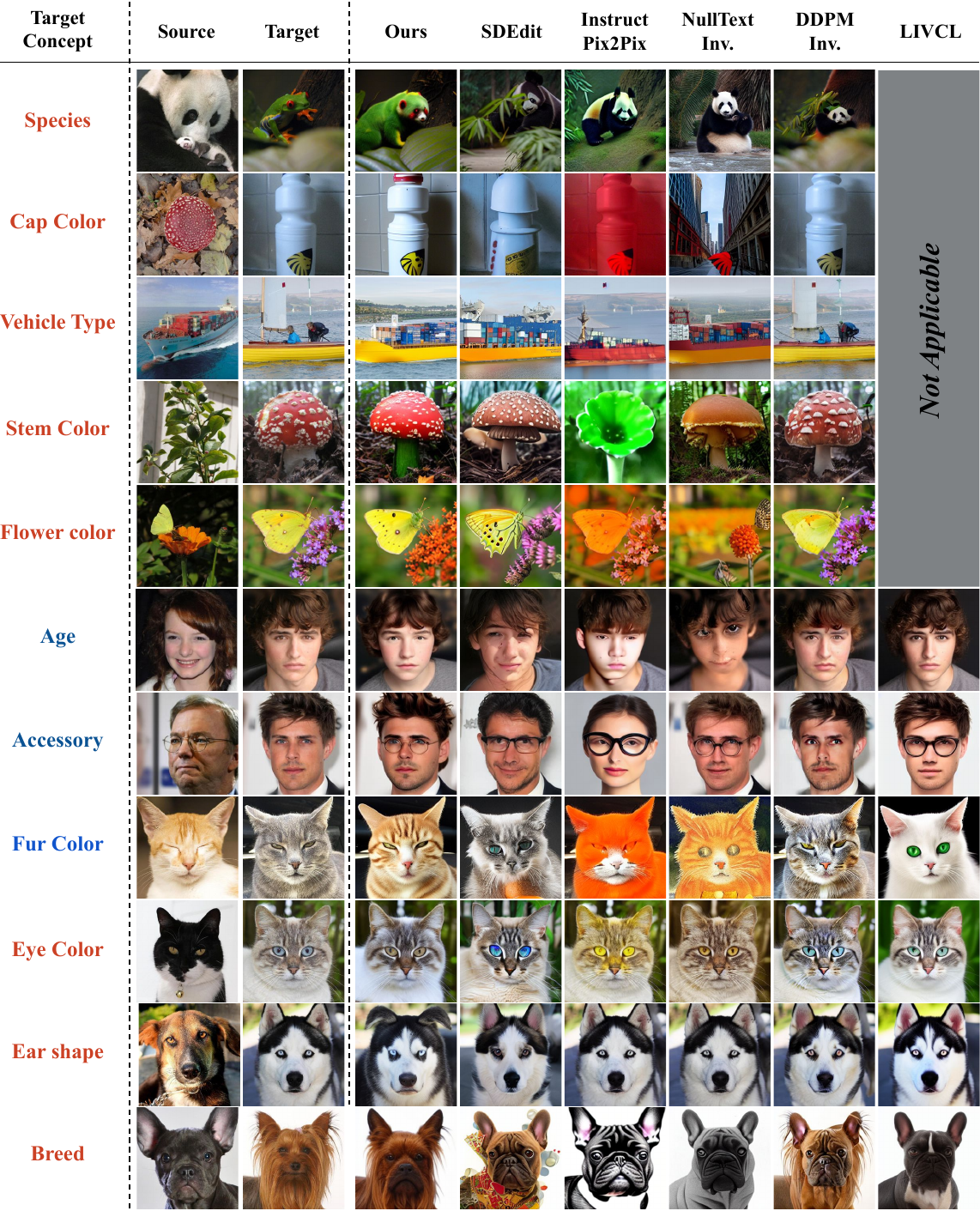}
\end{minipage}
\end{minipage}
\caption{Qualitative results on ImageNet-S20, CelebA-HQ and AFHQ datasets. Our framework grounds visual concepts to diverse concept axes in real-world images. Note that $\textbf{\color{BrickRed}{red\:\:concepts}}$ are not provided by our prompt but rather adaptively discovered by VLM. Since it's infeasible to predefine all axes covering the whole dataset like ImageNet-S20, LIVCL was not applicable for ImageNet-S20.
}
\label{fig:visual_concept_editing}
\vspace{-0.3in}
\end{figure*}

\paragraph{Dataset}
\vspace{-0.05in}
We validate our framework on complex and unstructured real-world data, where each image contains a diverse set of conceptual axes that is infeasible to manually predefine these axes to cover all possible variations within the data. 
To this end, we first conduct experiments on a subset of the ImageNet dataset. 
We randomly sampled 20 classes from ImageNet (referred to as ImageNet-S20), covering categories such as animals (\textit{e.g.}, tree frog, American black bear, sulphur butterfly, giant panda), everyday objects (\textit{e.g.}, padlock, grand piano, motor scooter), and scenes (\textit{e.g.}, boathouse, water tower), yielding approximately 28k training images ($\sim$1.4k images per class). 
This dataset presents a challenging scenario as each class contains diverse, image-specific visual concepts that are often not shared by other classes.
Given the infeasibility of manually defining all the concept axes in ImageNet-S20 for prior visual concept learning methods~\cite{lee2023livcl}, 
we additionally compare our approach with \citep{lee2023livcl} using relatively controlled datasets with diverse concept axes, such as CelebA-HQ~\cite{karras2017progressive}, AFHQ-Dog, and AFHQ-Cat~\cite{choi2020stargan2_afhq}. 
We collect the frequently observed axes discovered by our method per dataset and use them to train the baseline~\cite{lee2023livcl}
All images are resized to 256$\times$256 for our experiments. For training and validation, we use the following splits: 28k/0.6k images for ImageNet-S20, 27k/3k for CelebA-HQ, and around 5k/0.5k for AFHQ-Dog and AFHQ-Cat.

\vspace{-0.08in}
\paragraph{Evaluation Protocol}
To evaluate whether the concept representations faithfully capture their associated semantics and are disentangled from other axes, we perform a \textit{visual concept editing} task. 
In this task, we select source and target images and identify the concept axes to be edited. 
The objective is to transfer a visual concept from the source to the target image without affecting other attributes. 
For evaluation, we use the top-50 and top-10 most frequently discovered axes per dataset for ImageNet-S20 and the remaining datasets, respectively, excluding axes that remain constant across the dataset, such as the subject type in CelebA-HQ, which is always human.
For quantitative evaluation, we measure the CLIP-Score~\cite{radford2021clip} and BLIP-Score~\cite{li2022blip} between the edited images and their corresponding swapped text descriptions $V^c$. 
Specifically, we construct text prompts $V^c$ such as "a photo of a cat with {brown, fluffy, striped} fur, against a {black} background," and evaluate the alignment with the images using CLIP and BLIP. Additionally, we conduct human evaluation, collecting responses from 10 participants per dataset via Prolific~\cite{prolific}, following the procedure in \citet{lee2023livcl}. Details on the human evaluation setup are provided in Appendix~\ref{subsec:human_eval}.

\vspace{-0.08in}
\paragraph{Baselines}
We compare our method to LIVCL~\cite{lee2023livcl}, a recent visual concept learning approach that extracts concept representations along predefined primitive axes such as color, category, and style. 
As LIVCL explored in low resolution images, \textit{e.g.}, 64$\times$64 pixels, we replace its T2I decoder and pretrained image encoder with Stable Diffusion~\cite{rombach2022ldm} and DINO-
v2~\cite{oquab2023dinov2}, respectively. 
Since LIVCL requires predefined axes for training, we used the top-50/10 most frequent axes discovered by our method for ImageNet-S20 and the others, respectively.
We also compare our method to four recent text-based image editing methods—
SDEdit~\cite{meng2021sdedit}, InstructPix2Pix~\cite{brooks2023instructpix2pix}, Null-text Inversion~\cite{mokady2023nulltext},  and DDPM-Inversion~\cite{huberman2024ddpm_inversion}. 
Although these baselines lack mechanisms for extracting visual concepts from source images, we instead edit the image with GT text descriptions given by the VLM. For each method, we used a prompt including target attributes to be changed, such as "a photo of a dog with brown fur" for editing. We used default hyper-parameters for text-based editing methods. 

\vspace{-0.03in}
\subsection{Main Results}
\vspace{-0.08in}
\paragraph{Quantitative Results}

\setlength\intextsep{0pt}
\begin{wraptable}[9]{r}{0.43\textwidth}
  \centering
  \scriptsize
  \setlength{\tabcolsep}{2pt}
  \caption{Human evaluation results.}
  \vspace{-0.05in}
  \label{tab:human_eval}
    \begin{tabular}{@{}c|ccc@{}}
    \toprule
    Method             & CelebA-HQ      & AFHQ-Dog       & AFHQ-Cat       \\ \midrule
    SDEdit             & 0.448          & 0.486          & 0.464          \\
    InstructPix2Pix    & 0.465          & 0.385          & 0.416          \\
    NullText Inversion & 0.414          & 0.514          & 0.442          \\
    DDPM-Inversion     & 0.528          & 0.548          & 0.584          \\
    LIVCL              & 0.465          & 0.478          & 0.471          \\
    Ours               & \textbf{0.636} & \textbf{0.589} & \textbf{0.623} \\ \bottomrule
    \end{tabular}
  \vspace{-0.1in}
\end{wraptable}

We report quantitative comparison of our method to the baselines in Table~\ref{table:overall_quat_values}. 
Our methods consistently outperform all baselines on all of the datasets by a clear margin. 
High CLIP and BLIP scores demonstrate the effectiveness of our method in capturing image-related visual concepts. A human evaluation in Table~\ref{tab:human_eval} provides a more direct assessment of reflecting subtle visual nuances. Since text-based editing methods are inherently independent of source images and LIVCL struggles to encode image-dependent details due to its training objective, the performance gap becomes even more pronounced in the human evaluation. These results validate the effectiveness of our framework in visual grounding with diverse axes in real-world scenes. 

\paragraph{Qualitative Comparison}
\vspace{-0.1in}
Figure~\ref{fig:visual_concept_editing} presents the qualitative results on visual concept editing. 
Our method identified a diverse set of image-related axes and discovered novel concepts such as species, cap color, vehicle type, eye color, and breed, which were not specified in the prompt.
It demonstrates that our universal prompt can generalize to unseen domains.
Within the discovered axes, our method accurately alters each concept without affecting others.
In contrast, LIVCL often fails to encode image-specific details, such as generating different glasses in the seventh row, last column, or 
disentangling from other axes like changing fur color and texture in the eighth row, last column.
We attribute this to the inherent trade-off in LIVCL’s objective between concept disentanglement and capturing image-dependent details.
Thanks to our compositional anchoring objective, our method achieves both disentanglement along each axis and the preservation of image-specific details, \textit{e.g.}, transferring similar glasses in the seventh row of third column. 
Text-based approaches also struggle with concept-wise manipulation, often modifying the global color of images (InstructPix2Pix and NullText-Inversion) or leaving
them unchanged (SDEdit and DDPM-Inversion). 
Even when they transfer the correct attribute, they fail to capture the visual nuances of source attributes. 
For further visual inspection, please refer to additional qualitative results in Appendix~\ref{subsec:more_qualitative}.

\vspace{-0.1in}
\begin{figure*}[!t]
  \centering
  \begin{minipage}[c]{0.98\textwidth}
    \centering
    \scriptsize
    \includegraphics[width=0.9\linewidth]{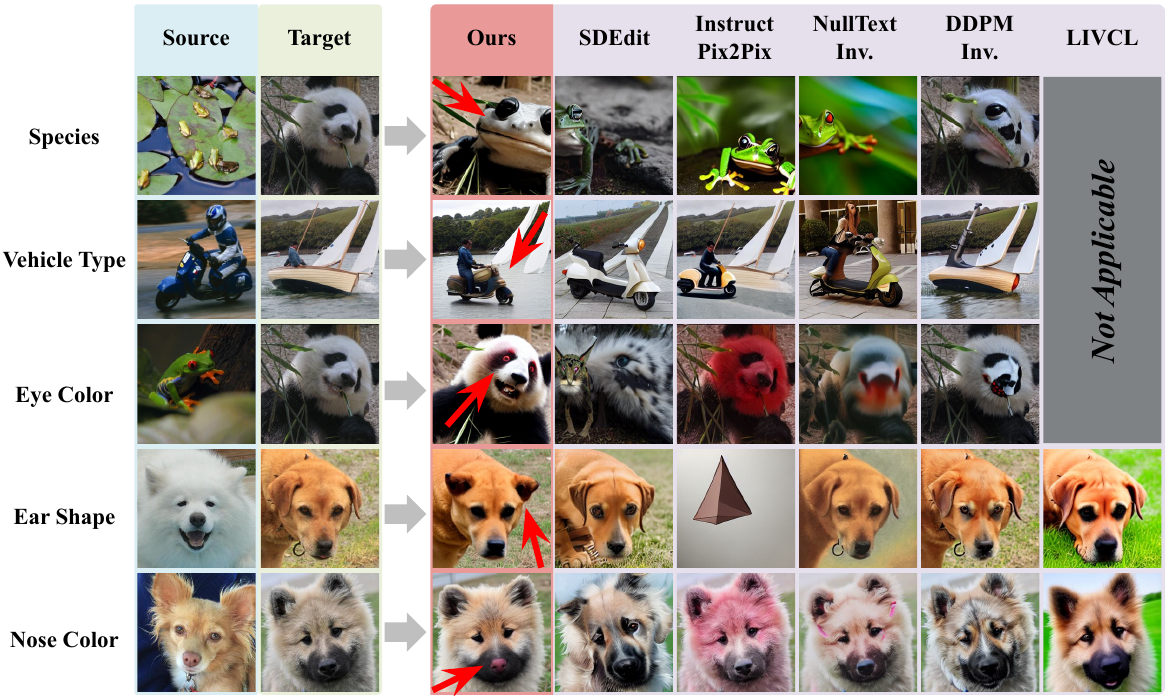}
    \captionof{figure}{Compositional generalization to unseen concept combination. Given OOD combination of concepts such as frog with panda's fur pattern, only our method generates plausible results.}
    \label{fig:qual_comp_gen}
  \end{minipage}
  \vspace{-0.2in}
\end{figure*}

\paragraph{Compositional Generalization}
Interestingly, our method demonstrates superior compositional generalization to unseen combinations of concepts compared to the baselines, as shown in Figure~\ref{fig:qual_comp_gen}. In the figure, our method successfully generates novel compositions, such as a large frog with a panda's fur pattern, pandas with red eyes, or scooters floating on water, which do not exist in the real world. 
In contrast, the baselines either alter multiple attributes simultaneously or change nothing at all, and fail to generate plausible generalizations. 
For instance, when modifying species or eye color, body colors are also changed as seen in the fifth and sixth columns of the first row, and the fifth column of the third row. Furthermore, while all baselines struggle with manipulating ear shapes or nose colors, which are strongly correlated with dog breed, our method shapes the ears of a Labrador into a triangle (third column of fourth row) and renders a dog’s nose in pink (third column of fifth row). 
We conjecture that such compositional generalization arises from our compositional anchoring objective, which explicitly promotes random composite images to exhibit corresponding compositions of attributes.  

\vspace{-0.03in}
\paragraph{Composition From Multiple Images}
\begin{figure*}[!t]
\centering
\begin{minipage}[!t]{\textwidth}
\scriptsize
\centering
\captionof{table}{
Comparisons of visual concept editing. Our method outperforms recent text-based editing methods~\cite{meng2021sdedit,mokady2023nulltext,brooks2023instructpix2pix,huberman2024ddpm_inversion} and language-informed visual concept learning~\cite{lee2023livcl}.  
} 
\vspace{-0.08in}
\label{table:multiple_composition}
\setlength{\tabcolsep}{2pt}
\begin{tabular}{@{}c|cccccc|cccccc|cccccc@{}}
\toprule
\multirow{3}{*}{method} & \multicolumn{6}{c|}{CelebA-HQ}                                                                                           & \multicolumn{6}{c|}{AFHQ-Dog}                                                                                            & \multicolumn{6}{c}{AFHQ-Cat}                                                                                             \\ \cmidrule(l){2-19} 
                        & \multicolumn{3}{c|}{CLIP}                                             & \multicolumn{3}{c|}{BLIP}                        & \multicolumn{3}{c|}{CLIP}                                             & \multicolumn{3}{c|}{BLIP}                        & \multicolumn{3}{c|}{CLIP}                                             & \multicolumn{3}{c}{BLIP}                         \\ \cmidrule(l){2-19} 
                        & N=2            & N=3            & \multicolumn{1}{c|}{N=4}            & N=2            & N=3            & N=4            & N=2            & N=3            & \multicolumn{1}{c|}{N=4}            & N=2            & N=3            & N=4            & N=2            & N=3            & \multicolumn{1}{c|}{N=4}            & N=2            & N=3            & N=4            \\ \midrule
SDEdit                  & 0.203          & 0.202          & \multicolumn{1}{c|}{0.204}          & 0.443          & 0.440          & 0.439          & 0.251          & 0.254          & \multicolumn{1}{c|}{0.255}          & 0.493          & 0.496          & 0.494          & 0.257          & 0.254          & \multicolumn{1}{c|}{0.254}          & 0.466          & 0.456          & 0.443          \\
InstructPix2Pix         & 0.201          & 0.199          & \multicolumn{1}{c|}{0.197}          & 0.416          & 0.413          & 0.411          & 0.225          & 0.224          & \multicolumn{1}{c|}{0.221}          & 0.456          & 0.458          & 0.453          & 0.248          & 0.250          & \multicolumn{1}{c|}{0.249}          & 0.462          & 0.463          & 0.450          \\
NullText Inv.      & 0.206          & 0.199          & \multicolumn{1}{c|}{0.194}          & 0.428          & 0.420          & 0.417          & 0.247          & 0.250              & \multicolumn{1}{c|}{0.250}          & 0.479              & 0.482          & 0.480          & 0.256          & 0.257          & \multicolumn{1}{c|}{0.258}          & 0.471          & 0.469          & 0.468          \\
DDPM Inv.          & 0.213          & 0.207          & \multicolumn{1}{c|}{0.203}          & 0.463          & 0.449          & 0.437          & 0.262          & 0.260          & \multicolumn{1}{c|}{0.257}          & 0.506          & 0.501          & 0.493          & 0.261          & 0.258          & \multicolumn{1}{c|}{0.253}          & 0.473          & 0.458          & 0.440          \\
LIVCL                   & 0.225          & 0.219          & \multicolumn{1}{c|}{0.214}          & 0.454          & 0.440          & 0.429          & 0.267          & 0.264          & \multicolumn{1}{c|}{0.260}          & 0.507          & 0.502          & 0.491          & 0.262          & 0.257          & \multicolumn{1}{c|}{0.250}          & 0.463          & 0.447          & 0.428          \\
Ours                    & \textbf{0.238} & \textbf{0.236} & \multicolumn{1}{c|}{\textbf{0.236}} & \textbf{0.492} & \textbf{0.490} & \textbf{0.491} & \textbf{0.269} & \textbf{0.266} & \multicolumn{1}{c|}{\textbf{0.262}} & \textbf{0.528} & \textbf{0.528} & \textbf{0.523} & \textbf{0.271} & \textbf{0.269} & \multicolumn{1}{c|}{\textbf{0.268}} & \textbf{0.516} & \textbf{0.513} & \textbf{0.512} \\ \bottomrule
\end{tabular}

\end{minipage}
\begin{minipage}[!t]{\textwidth}
\scriptsize
\centering
\begin{minipage}[!t]{1.0\textwidth}
\scriptsize
\centering
\includegraphics[width=0.95\linewidth]{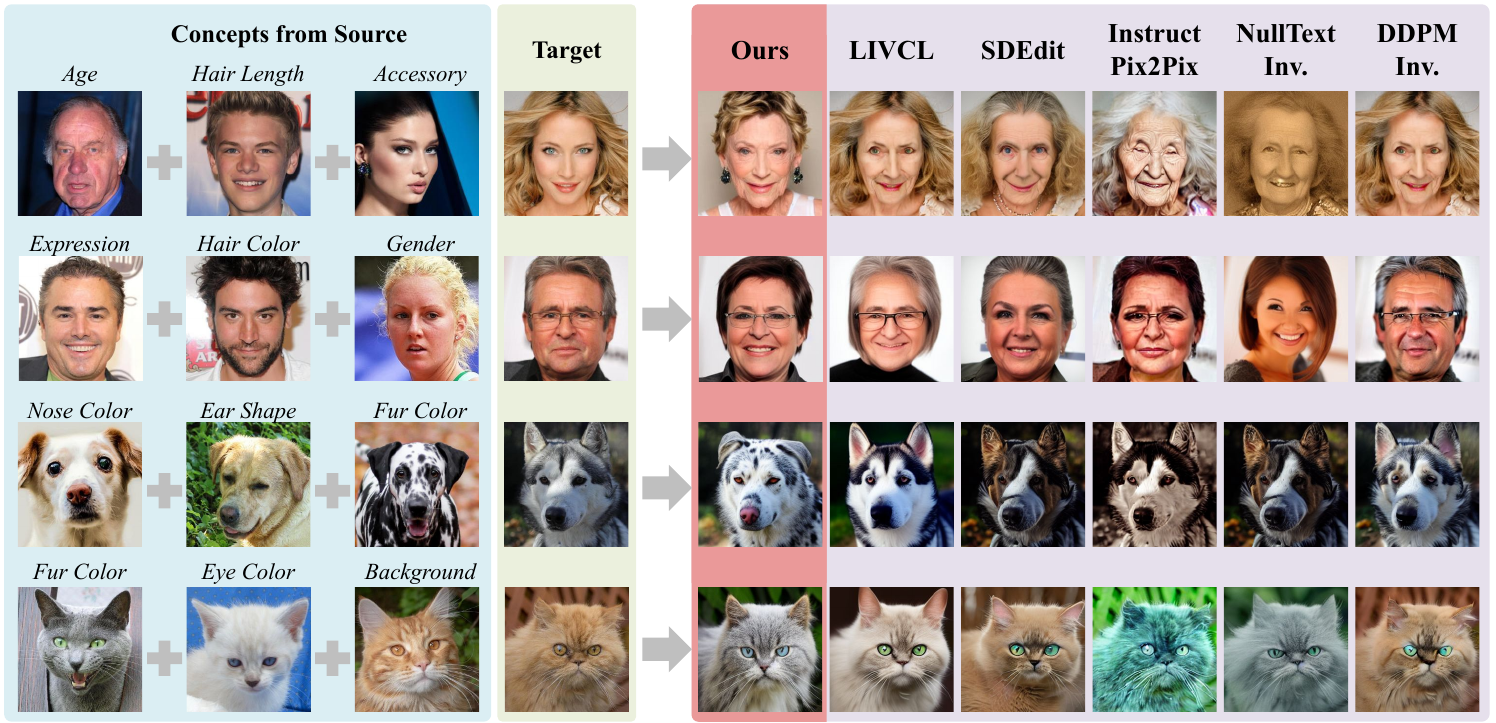}
\end{minipage}
\end{minipage}
\vspace{-0.05in}
\caption{Composition of visual concepts from multiple images. Only our method accurately reflects all of the attributes of source images.}
\label{fig:qual_multiple_comp}

\begin{minipage}[!t]{1.0\textwidth}
\scriptsize
\centering
\includegraphics[width=0.95\linewidth]{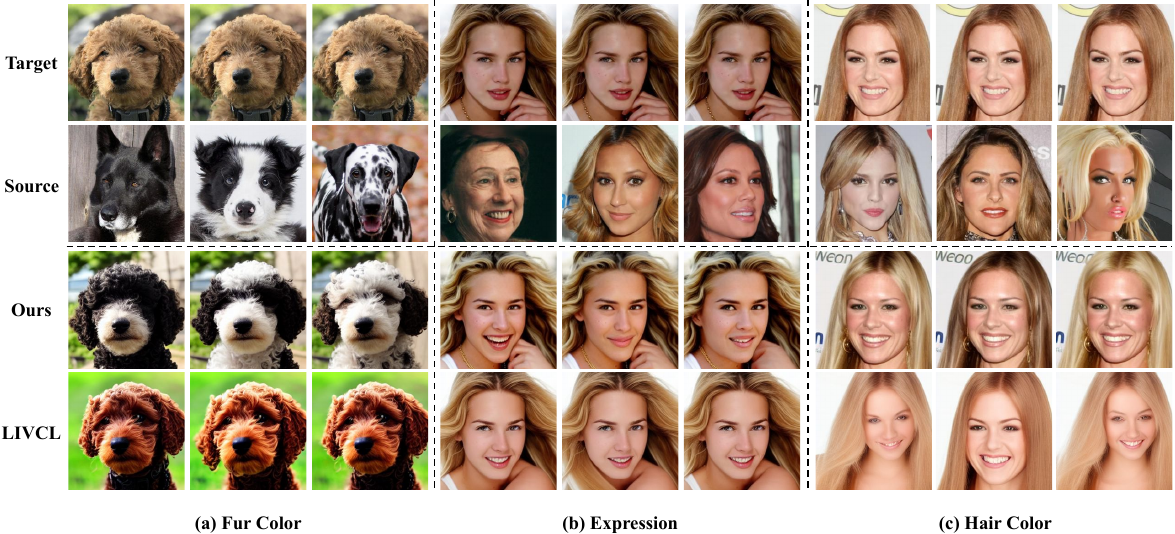}
\vspace{-0.05in}
\captionof{figure}{Examples of visual nuance transfer. Even when transferring the same attributes, \textit{e.g.}, black and white fur or blonde hair, the outputs reflect subtle details of source images.}
\label{fig:qual_visual_nuance}
\end{minipage}
\vspace{-0.15in}

\end{figure*}
\vspace{-0.1in}
To further analyze the quality of extracted visual concepts, we consider the more challenging multi-image composition task. For each target image, we select $N$ source images and randomly sample unique concept axes from each source, \textit{i.e.}, $N$ different axes. We then edit the target image along those axes to produce composite images. 
We conduct this task only on CelebA-HQ and AFHQ, as the high image diversity within ImageNet classes (\textit{e.g.}, partial views, different viewpoints, or varying light conditions) often leads to cases where the concept axes are not consistently shared among the same class images, resulting in noisy evaluations. In contrast, CelebA-HQ and AFHQ have more controlled structures, making them better suited for this task.

Table \ref{table:multiple_composition} presents CLIP and BLIP scores for editing up to four axes. Our method again consistently outperforms all baselines. Moreover, all baselines suffer a clear drop in both metrics as $N$ increases, whereas our method shows only a marginal decrease. This robustness indicates that our concept representations are well disentangled and faithfully capture the correct semantics of the input images. 
Figure \ref{fig:qual_multiple_comp} shows qualitative results for $N=3$. 
Composite images from our method are faithfully modified to reflect all of the source images' concepts. In contrast, the baseline models often omit or distort certain attributes. For example, all baselines fail to render the short hairstyle and blue earrings (first row), and some either drop the facial expression (InstructPix2Pix, DDPM Inversion) or misapply the hair color (LIVCL, SDEdit) in the composite outputs (second row).

\vspace{-0.1in}
\paragraph{Visual Nuance Transfer}
In contrast to text-based editing methods, visual concept learning methods can capture visual nuances in the continuous representation space. 
Since LIVCL is not applicable to ImageNet-S20, we compare our methods to LIVCL on the CelebA-HQ and AFHQ datasets. 
Figure~\ref{fig:qual_visual_nuance} highlights visual nuances captured in the concept representation of our method. 
In the figure, our method transfers subtle visual details such as detailed fur patterns, subtle differences in smiles, or hair color tones. 
In contrast, LIVCL struggles to correctly transfer these visual details, \textit{e.g.}, 
the resulting image always exhibits the same expressions in Figure~\ref{fig:qual_visual_nuance}(b). 
It even fails to reconstruct the original images in Figure~\ref{fig:qual_visual_nuance}(c). 
This implies the suboptimal trade-off between concept disentanglement and 
image-dependent encoding induced by the objective in LIVCL. 
While pushing concept representations $\mathbf{z}_i$ closer to text embeddings $\mathbf{v}_i$, \textit{i.e.}, $\mathbf{z}_i=\mathbf{v}_i$, can achieve disentanglement, it sacrifices  visual information.
In contrast, our compositional anchoring objective bypasses such a trade-off and thereby achieves both disentanglement and rich image-dependent details within representations. 
More qualitative results on visual nuance transfer are provided in Appendix~\ref{subsec:more_qualitative}.  

\vspace{-0.1in}
\subsection{Ablation Study}
\vspace{-0.1in}
In this section, we conduct an ablation study on VLM choices, architectural design choices, and objective functions to examine the robustness and effectiveness of our choices. All of the experiments are evaluated on the visual concept editing task in the CelebA-HQ dataset.
\vspace{-0.1in}
\paragraph{VLM choices}

Since the discovery of concept axes in our method is directly affected by the quality of VLM outputs, we examine the robustness of our framework on two additional popular open-sourced VLMs (Qwen2.5-VL\cite{Qwen2.5-VL}, Ovis2\cite{ovis2}), which have ranked highly on reasoning benchmarks. 
Moreover, as it is difficult to directly control or quantify VLM performance, we instead control output quality by dropping partial axes (e.g., 10\% and 20\%) from the VLM outputs. It is a practical scenario as VLMs cannot always capture the complete axes for a given scene. 
Table~\ref{tab:vlm_choice} shows that our method is robust to both VLM choices and missing axes. 
We hypothesize that this is because even though some image-related axes can be missed in each example, those axes will eventually be repeatedly exposed across the dataset. 
Additionally, since our compositional anchoring loss encourages the compositionality of the concept representations, our framework might be internally trained for better compositional generalization, which improves adaptation with fewer samples. 
In fact, our method is capable of generating OOD samples (Figure~\ref{fig:qual_comp_gen}). 
The robustness of our framework regarding the performance of VLMs suggests that it can scale to more complex real-world datasets, as VLMs do not always need to capture complete axes for every scene.

\begin{table*}[t]

\centering
\setlength{\tabcolsep}{3pt}
\renewcommand{\arraystretch}{0.9}

\begin{minipage}{0.41\textwidth}
\centering
\scriptsize
\caption{Ablation study on VLM choices.}
\vspace{-0.05in}
\label{tab:vlm_choice}
\begin{tabular}{@{}lcc@{}}
\toprule
VLM choices            & CLIP & BLIP \\ \midrule
Qwen2.5-VL~\cite{Qwen2.5-VL}             & 23.72 & 48.64 \\
Ovis2~\cite{ovis2}                  & 23.48 & 48.35 \\
InternVL2-5 (Ours)            & \textbf{23.88} & \textbf{49.58} \\
InternVL2-5 + 10\% drop & 23.52 & 48.69 \\
InternVL2-5 + 20\% drop & 23.65 & 48.61 \\ 
\bottomrule
\end{tabular}
\vspace{-0.05in}
\end{minipage}
\hfill
\begin{minipage}{0.58\textwidth}
\centering
\scriptsize
\caption{Ablation study on Architectural choices.}
\vspace{-0.05in}
\label{tab:architecture_choice}

\begin{tabular}{@{}ccc|cc@{}}
\toprule
\multicolumn{3}{c|}{Archiectural choices}                                                  & \multicolumn{2}{c}{Metrics}                                                                                                   \\ \midrule
{ Decoder} & Vision Encoder & Concept Encoder & { CLIP}           & { BLIP}           \\ \midrule
Frozen T2I                                             & Dinov2         & UCE             & { \textbf{23.88}} & { 49.58}          \\
LoRA-finetuned T2I                                     & Dinov2         & UCE             & { 23.67}          & { \textbf{49.62}} \\
Frozen T2I                                             & CLIP           & UCE             & { 22.03}          & { 46.21}          \\
Frozen T2I                                             & Dinov2         & Shared MLP      & { 21.63}          & { 46.8}           \\ \bottomrule
\end{tabular}

\vspace{-0.1in}
\end{minipage}
\vspace{-0.1in}
\end{table*}

\vspace{-0.05in}
\paragraph{Architectural choices}

Table~\ref{tab:architecture_choice} presents the ablation studies on architectural choices as follows: (1) Frozen T2I decoder versus LoRA-finetuned decoder, (2) choice of vision encoder (Dinov2 versus CLIP), and (3) universal concept encoder (UCE in the Table~\ref{tab:architecture_choice}) versus shared MLP architectures. 
First, finetuning the decoder with LoRA does not affect overall performance. 
Large-scale pretrained T2I models have already learned expressive data priors on natural images, facilitating faster training of the generation model. 
Therefore, the frozen T2I model does not bottleneck our framework. 
Replacing the Dinov2 encoder with the CLIP encoder causes a significant performance drop, as CLIP is trained for text-alignment, making its discriminative properties inferior to those of recent self-supervised methods like Dinov2.
Lastly, replacing our universal concept encoder with a shared MLP architecture, which is a naive version of an axis-agnostic encoder, also results in a severe drop.
Specifically, the visual feature is mean-pooled into a vector, concatenated with axis embeddings, and passed through shared MLP layers to encode concept representations. 
To make this encoder generally work for diverse concept axes, we shared this MLP layer for all of the axes. 
This approach likely fails because the shared MLP treats each concept independently, blocking complex interactions between concepts. It clearly highlights the effectiveness of our universal concept encoder.

\vspace{-0.1in}
\paragraph{Component-wise Contribution}
\setlength\intextsep{0pt}
\begin{wraptable}[8]{r}{0.35\textwidth}
  \centering
  \tiny
  \caption{Ablation study on our method. Both $\mathcal{L}_{\mathrm{Comp}}$ and $g_\phi$ contribute to concept disentanglement.}
  \label{tab:ablation_study}
  \begin{tabular}{cc|cc}
    \toprule
    $\mathcal{L}_{\mathrm{Comp}}$ & $g_\phi$     & CLIP ($\uparrow$) & BLIP ($\uparrow$) \\ 
    \midrule
    \checkmark & \xmark      & 21.1              & 44.72             \\
    \xmark     & \checkmark  & 22.89             & 47.47             \\
    \checkmark & \checkmark  & \textbf{23.88}    & \textbf{49.58}    \\
    \bottomrule
  \end{tabular}
  \vspace{-0.1in}
\end{wraptable}

We conduct an ablation study on each component in our objective for concept disentanglement, \textit{i.e.}, $g_\phi$ and $\mathcal{L}_{\text{Comp}}$, and report the results in Table~\ref{tab:ablation_study}. 
Without employing $g_\phi$ and instead directly regressing each concept representation $\mathbf{z}_i$ to $\mathbf{v}_i$ in Equation~\ref{eqn:comp_anchor_loss} and \ref{eqn:regression_loss}, 
we observe significant drops in both CLIP-Score and BLIP-Score. 
This result indicates the importance of $g_\phi$ in preventing a direct trade-off between 
disentanglement and encoding image-dependent details.
Furthermore, removing $\mathcal{L}_{\text{Comp}}$ also leads to suboptimal CLIP- and BLIP-Score. 
This is because minimizing Equation~\ref{eqn:regression_loss} only guarantees $\mathbf{z}_i$ to have information of $\mathbf{v}_i$ 
but does not prevent it from encoding entangled information related to other concept axes.  
\vspace{-0.1in}
\section{Conclusion}
\vspace{-0.1in}
In this study, we present a scalable framework for grounding visual concepts along adaptive concept axes in real-world scenes. 
Our framework leverages a pretrained VLM and universal prompt design 
to adaptively identify diverse, image-related concept axes. 
A single, unified concept encoder then binds visual features to these axes, eliminating the need for separate per-concept encoders. 
To ensure each axis remains disentangled while preserving instance-level detail, we introduce a compositional anchoring loss. We randomly swap concept representations across images and regularize the resulting composite outputs to match their corresponding text descriptions. 
In the visual concept editing task on real-world datasets, our method 
consistently outperforms prior approaches in language-informed visual concept learning and recent text-based editing methods, demonstrating the effectiveness of our framework in learning adaptive visual concepts in real-world datasets. Also, our approach demonstrates successful transfer of subtle visual nuances and stronger compositional generalization.

\paragraph{Acknowledgment}
This work was in part supported by the National Research Foundation of Korea (RS-2024-00351212 and RS-2024-00436165) and the Institute of Information \& communications Technology Planning \& Evaluation (IITP) (RS-2022-II220926, RS-2022-II220959, and RS-2024-00509279) funded by the Korea government (MSIT).

\bibliographystyle{plainnat}  

\bibliography{main}

\begin{thebibliography}{38}
\providecommand{\natexlab}[1]{#1}
\providecommand{\url}[1]{\texttt{#1}}
\expandafter\ifx\csname urlstyle\endcsname\relax
  \providecommand{\doi}[1]{doi: #1}\else
  \providecommand{\doi}{doi: \begingroup \urlstyle{rm}\Url}\fi

\bibitem[Anderson et~al.(2018)Anderson, He, Buehler, Teney, Johnson, Gould, and Zhang]{anderson2018bottom}
Peter Anderson, Xiaodong He, Chris Buehler, Damien Teney, Mark Johnson, Stephen Gould, and Lei Zhang.
\newblock Bottom-up and top-down attention for image captioning and visual question answering.
\newblock In \emph{Proceedings of the IEEE conference on computer vision and pattern recognition}, pages 6077--6086, 2018.

\bibitem[Bai et~al.(2025)Bai, Chen, Liu, Wang, Ge, Song, Dang, Wang, Wang, Tang, Zhong, Zhu, Yang, Li, Wan, Wang, Ding, Fu, Xu, Ye, Zhang, Xie, Cheng, Zhang, Yang, Xu, and Lin]{Qwen2.5-VL}
Shuai Bai, Keqin Chen, Xuejing Liu, Jialin Wang, Wenbin Ge, Sibo Song, Kai Dang, Peng Wang, Shijie Wang, Jun Tang, Humen Zhong, Yuanzhi Zhu, Mingkun Yang, Zhaohai Li, Jianqiang Wan, Pengfei Wang, Wei Ding, Zheren Fu, Yiheng Xu, Jiabo Ye, Xi~Zhang, Tianbao Xie, Zesen Cheng, Hang Zhang, Zhibo Yang, Haiyang Xu, and Junyang Lin.
\newblock Qwen2.5-vl technical report.
\newblock \emph{arXiv preprint arXiv:2502.13923}, 2025.

\bibitem[Brooks et~al.(2023)Brooks, Holynski, and Efros]{brooks2023instructpix2pix}
Tim Brooks, Aleksander Holynski, and Alexei~A Efros.
\newblock Instructpix2pix: Learning to follow image editing instructions.
\newblock In \emph{Proceedings of the IEEE/CVF conference on computer vision and pattern recognition}, pages 18392--18402, 2023.

\bibitem[Chen et~al.(2024)Chen, Wu, Wang, Su, Chen, Xing, Zhong, Zhang, Zhu, Lu, et~al.]{chen2024internvl}
Zhe Chen, Jiannan Wu, Wenhai Wang, Weijie Su, Guo Chen, Sen Xing, Muyan Zhong, Qinglong Zhang, Xizhou Zhu, Lewei Lu, et~al.
\newblock Internvl: Scaling up vision foundation models and aligning for generic visual-linguistic tasks.
\newblock In \emph{Proceedings of the IEEE/CVF conference on computer vision and pattern recognition}, pages 24185--24198, 2024.

\bibitem[Choi et~al.(2020)Choi, Uh, Yoo, and Ha]{choi2020stargan2_afhq}
Yunjey Choi, Youngjung Uh, Jaejun Yoo, and Jung-Woo Ha.
\newblock Stargan v2: Diverse image synthesis for multiple domains.
\newblock In \emph{Proceedings of the IEEE/CVF conference on computer vision and pattern recognition}, pages 8188--8197, 2020.

\bibitem[Clark et~al.(2023)Clark, Vicol, Swersky, and Fleet]{clark2023draft}
Kevin Clark, Paul Vicol, Kevin Swersky, and David~J Fleet.
\newblock Directly fine-tuning diffusion models on differentiable rewards.
\newblock \emph{arXiv preprint arXiv:2309.17400}, 2023.

\bibitem[Everingham et~al.(2010)Everingham, Van~Gool, Williams, Winn, and Zisserman]{everingham2010pascal}
Mark Everingham, Luc Van~Gool, Christopher~KI Williams, John Winn, and Andrew Zisserman.
\newblock The pascal visual object classes (voc) challenge.
\newblock \emph{International journal of computer vision}, 88:\penalty0 303--338, 2010.

\bibitem[Farhadi et~al.(2009)Farhadi, Endres, Hoiem, and Forsyth]{farhadi2009describe_by_attr}
Ali Farhadi, Ian Endres, Derek Hoiem, and David Forsyth.
\newblock Describing objects by their attributes.
\newblock In \emph{2009 IEEE conference on computer vision and pattern recognition}, pages 1778--1785. IEEE, 2009.

\bibitem[Gai et~al.(2021)Gai, Jain, Zhang, Gonzalez, Song, and Stoica]{gai-etal-2021-grounded-graph}
Yu~Gai, Paras Jain, Wendi Zhang, Joseph Gonzalez, Dawn Song, and Ion Stoica.
\newblock Grounded graph decoding improves compositional generalization in question answering.
\newblock In Marie-Francine Moens, Xuanjing Huang, Lucia Specia, and Scott Wen-tau Yih, editors, \emph{Findings of the Association for Computational Linguistics: EMNLP 2021}, pages 1829--1838, Punta Cana, Dominican Republic, November 2021. Association for Computational Linguistics.
\newblock \doi{10.18653/v1/2021.findings-emnlp.157}.
\newblock URL \url{https://aclanthology.org/2021.findings-emnlp.157/}.

\bibitem[Gal et~al.(2022)Gal, Alaluf, Atzmon, Patashnik, Bermano, Chechik, and Cohen-Or]{gal2022image}
Rinon Gal, Yuval Alaluf, Yuval Atzmon, Or~Patashnik, Amit~H Bermano, Gal Chechik, and Daniel Cohen-Or.
\newblock An image is worth one word: Personalizing text-to-image generation using textual inversion.
\newblock \emph{arXiv preprint arXiv:2208.01618}, 2022.

\bibitem[Huberman-Spiegelglas et~al.(2024)Huberman-Spiegelglas, Kulikov, and Michaeli]{huberman2024ddpm_inversion}
Inbar Huberman-Spiegelglas, Vladimir Kulikov, and Tomer Michaeli.
\newblock An edit friendly ddpm noise space: Inversion and manipulations.
\newblock In \emph{Proceedings of the IEEE/CVF Conference on Computer Vision and Pattern Recognition}, pages 12469--12478, 2024.

\bibitem[Jung et~al.(2024)Jung, Yoo, Ahn, and Hong]{jung2024l2c}
Whie Jung, Jaehoon Yoo, Sungjin Ahn, and Seunghoon Hong.
\newblock Learning to compose: Improving object centric learning by injecting compositionality.
\newblock \emph{arXiv preprint arXiv:2405.00646}, 2024.

\bibitem[Karras et~al.(2017)Karras, Aila, Laine, and Lehtinen]{karras2017progressive}
Tero Karras, Timo Aila, Samuli Laine, and Jaakko Lehtinen.
\newblock Progressive growing of gans for improved quality, stability, and variation.
\newblock \emph{arXiv preprint arXiv:1710.10196}, 2017.

\bibitem[Koh et~al.(2020)Koh, Nguyen, Tang, Mussmann, Pierson, Kim, and Liang]{koh2020concept}
Pang~Wei Koh, Thao Nguyen, Yew~Siang Tang, Stephen Mussmann, Emma Pierson, Been Kim, and Percy Liang.
\newblock Concept bottleneck models.
\newblock In \emph{International conference on machine learning}, pages 5338--5348. PMLR, 2020.

\bibitem[Krizhevsky et~al.(2017)Krizhevsky, Sutskever, and Hinton]{krizhevsky2017imagenet}
Alex Krizhevsky, Ilya Sutskever, and Geoffrey~E Hinton.
\newblock Imagenet classification with deep convolutional neural networks.
\newblock \emph{Communications of the ACM}, 60\penalty0 (6):\penalty0 84--90, 2017.

\bibitem[Lee et~al.(2024)Lee, Zhang, Wu, and Wu]{lee2023livcl}
Sharon Lee, Yunzhi Zhang, Shangzhe Wu, and Jiajun Wu.
\newblock Language-informed visual concept learning.
\newblock \emph{ICLR}, 2024.

\bibitem[Li et~al.(2022)Li, Li, Xiong, and Hoi]{li2022blip}
Junnan Li, Dongxu Li, Caiming Xiong, and Steven Hoi.
\newblock Blip: Bootstrapping language-image pre-training for unified vision-language understanding and generation.
\newblock In \emph{International conference on machine learning}, pages 12888--12900. PMLR, 2022.

\bibitem[Li et~al.(2023)Li, Li, Savarese, and Hoi]{li2023blip-2}
Junnan Li, Dongxu Li, Silvio Savarese, and Steven Hoi.
\newblock Blip-2: Bootstrapping language-image pre-training with frozen image encoders and large language models.
\newblock In \emph{International conference on machine learning}, pages 19730--19742. PMLR, 2023.

\bibitem[Li et~al.(2020)Li, Huang, Hong, and Zhu]{li2020competence}
Qing Li, Siyuan Huang, Yining Hong, and Song-Chun Zhu.
\newblock A competence-aware curriculum for visual concepts learning via question answering.
\newblock In \emph{European Conference on Computer Vision}, pages 141--157. Springer, 2020.

\bibitem[Lin et~al.(2014)Lin, Maire, Belongie, Hays, Perona, Ramanan, Doll{\'a}r, and Zitnick]{lin2014mscoco}
Tsung-Yi Lin, Michael Maire, Serge Belongie, James Hays, Pietro Perona, Deva Ramanan, Piotr Doll{\'a}r, and C~Lawrence Zitnick.
\newblock Microsoft coco: Common objects in context.
\newblock In \emph{Computer vision--ECCV 2014: 13th European conference, zurich, Switzerland, September 6-12, 2014, proceedings, part v 13}, pages 740--755. Springer, 2014.

\bibitem[Lu et~al.(2024)Lu, Li, Chen, Xu, Luo, Zhang, and Ye]{ovis2}
Shiyin Lu, Yang Li, Qing-Guo Chen, Zhao Xu, Weihua Luo, Kaifu Zhang, and Han-Jia Ye.
\newblock Ovis: Structural embedding alignment for multimodal large language model.
\newblock \emph{arXiv:2405.20797}, 2024.

\bibitem[Mao et~al.(2019)Mao, Gan, Kohli, Tenenbaum, and Wu]{mao2019neuro}
Jiayuan Mao, Chuang Gan, Pushmeet Kohli, Joshua~B Tenenbaum, and Jiajun Wu.
\newblock The neuro-symbolic concept learner: Interpreting scenes, words, and sentences from natural supervision.
\newblock \emph{arXiv preprint arXiv:1904.12584}, 2019.

\bibitem[Meng et~al.(2021)Meng, He, Song, Song, Wu, Zhu, and Ermon]{meng2021sdedit}
Chenlin Meng, Yutong He, Yang Song, Jiaming Song, Jiajun Wu, Jun-Yan Zhu, and Stefano Ermon.
\newblock Sdedit: Guided image synthesis and editing with stochastic differential equations.
\newblock \emph{arXiv preprint arXiv:2108.01073}, 2021.

\bibitem[Mokady et~al.(2023)Mokady, Hertz, Aberman, Pritch, and Cohen-Or]{mokady2023nulltext}
Ron Mokady, Amir Hertz, Kfir Aberman, Yael Pritch, and Daniel Cohen-Or.
\newblock Null-text inversion for editing real images using guided diffusion models.
\newblock In \emph{Proceedings of the IEEE/CVF conference on computer vision and pattern recognition}, pages 6038--6047, 2023.

\bibitem[Nagarajan and Grauman(2018)]{nagarajan2018attributes_as_operator}
Tushar Nagarajan and Kristen Grauman.
\newblock Attributes as operators: factorizing unseen attribute-object compositions.
\newblock In \emph{Proceedings of the European Conference on Computer Vision (ECCV)}, pages 169--185, 2018.

\bibitem[Oquab et~al.(2023)Oquab, Darcet, Moutakanni, Vo, Szafraniec, Khalidov, Fernandez, Haziza, Massa, El-Nouby, et~al.]{oquab2023dinov2}
Maxime Oquab, Timoth{\'e}e Darcet, Th{\'e}o Moutakanni, Huy Vo, Marc Szafraniec, Vasil Khalidov, Pierre Fernandez, Daniel Haziza, Francisco Massa, Alaaeldin El-Nouby, et~al.
\newblock Dinov2: Learning robust visual features without supervision.
\newblock \emph{arXiv preprint arXiv:2304.07193}, 2023.

\bibitem[Prabhudesai et~al.(2023)Prabhudesai, Goyal, Pathak, and Fragkiadaki]{prabhudesai2023aligning}
Mihir Prabhudesai, Anirudh Goyal, Deepak Pathak, and Katerina Fragkiadaki.
\newblock Aligning text-to-image diffusion models with reward backpropagation, 2023.

\bibitem[Prolific(2014)]{prolific}
Prolific.
\newblock Prolific (version march 2025) [web platform], 2014.
\newblock First released in 2014. Copyright 2024. London, UK. Available at \url{https://www.prolific.com}.

\bibitem[Radford et~al.(2021)Radford, Kim, Hallacy, Ramesh, Goh, Agarwal, Sastry, Askell, Mishkin, Clark, et~al.]{radford2021clip}
Alec Radford, Jong~Wook Kim, Chris Hallacy, Aditya Ramesh, Gabriel Goh, Sandhini Agarwal, Girish Sastry, Amanda Askell, Pamela Mishkin, Jack Clark, et~al.
\newblock Learning transferable visual models from natural language supervision.
\newblock In \emph{International conference on machine learning}, pages 8748--8763. PmLR, 2021.

\bibitem[Rombach et~al.(2022)Rombach, Blattmann, Lorenz, Esser, and Ommer]{rombach2022ldm}
Robin Rombach, Andreas Blattmann, Dominik Lorenz, Patrick Esser, and Bj{\"o}rn Ommer.
\newblock High-resolution image synthesis with latent diffusion models.
\newblock In \emph{Proceedings of the IEEE/CVF conference on computer vision and pattern recognition}, pages 10684--10695, 2022.

\bibitem[Russakovsky et~al.(2015)Russakovsky, Deng, Su, Krause, Satheesh, Ma, Huang, Karpathy, Khosla, Bernstein, et~al.]{russakovsky2015imagenet}
Olga Russakovsky, Jia Deng, Hao Su, Jonathan Krause, Sanjeev Satheesh, Sean Ma, Zhiheng Huang, Andrej Karpathy, Aditya Khosla, Michael Bernstein, et~al.
\newblock Imagenet large scale visual recognition challenge.
\newblock \emph{International journal of computer vision}, 115:\penalty0 211--252, 2015.

\bibitem[Saari(2023)]{saari2023selecting}
Donald~G Saari.
\newblock Selecting a voting method: the case for the borda count.
\newblock \emph{Constitutional Political Economy}, 34\penalty0 (3):\penalty0 357--366, 2023.

\bibitem[Song et~al.(2020)Song, Meng, and Ermon]{song2020ddim}
Jiaming Song, Chenlin Meng, and Stefano Ermon.
\newblock Denoising diffusion implicit models.
\newblock \emph{arXiv preprint arXiv:2010.02502}, 2020.

\bibitem[Vinyals et~al.(2015)Vinyals, Toshev, Bengio, and Erhan]{vinyals2015show}
Oriol Vinyals, Alexander Toshev, Samy Bengio, and Dumitru Erhan.
\newblock Show and tell: A neural image caption generator.
\newblock In \emph{Proceedings of the IEEE conference on computer vision and pattern recognition}, pages 3156--3164, 2015.

\bibitem[Wiedemer et~al.(2023)Wiedemer, Brady, Panfilov, Juhos, Bethge, and Brendel]{wiedemer2023provable_ocl}
Thadd{\"a}us Wiedemer, Jack Brady, Alexander Panfilov, Attila Juhos, Matthias Bethge, and Wieland Brendel.
\newblock Provable compositional generalization for object-centric learning.
\newblock \emph{arXiv preprint arXiv:2310.05327}, 2023.

\bibitem[Zang et~al.(2024)Zang, Yun, Tan, Bui, and Sun]{zang2024vlm_discoverable_concept}
Yuan Zang, Tian Yun, Hao Tan, Trung Bui, and Chen Sun.
\newblock Pre-trained vision-language models learn discoverable visual concepts.
\newblock \emph{arXiv preprint arXiv:2404.12652}, 2024.

\bibitem[Zhang et~al.(2021)Zhang, Hu, Qiu, Shaw, and Sha]{zhang-etal-2021-visually-grounded}
Bowen Zhang, Hexiang Hu, Linlu Qiu, Peter Shaw, and Fei Sha.
\newblock Visually grounded concept composition.
\newblock In Marie-Francine Moens, Xuanjing Huang, Lucia Specia, and Scott Wen-tau Yih, editors, \emph{Findings of the Association for Computational Linguistics: EMNLP 2021}, pages 201--215, Punta Cana, Dominican Republic, November 2021. Association for Computational Linguistics.
\newblock \doi{10.18653/v1/2021.findings-emnlp.20}.
\newblock URL \url{https://aclanthology.org/2021.findings-emnlp.20/}.

\bibitem[Zhong et~al.(2024)Zhong, Hu, Lyu, and Wang]{zhong2024vis_table}
Yiwu Zhong, Zi-Yuan Hu, Michael~R Lyu, and Liwei Wang.
\newblock Beyond embeddings: The promise of visual table in multi-modal models.
\newblock \emph{arXiv preprint arXiv:2403.18252}, 2024.

\end{thebibliography}
\appendix

\newpage
\appendix
\onecolumn

\begin{appendix}
\setcounter{page}{1}
\section{Appendix}
\subsection{Limitations and Future Work}

In our work, as in most previous approaches, we cannot guarantee recovery of every ground-truth factor of variation. Some subtle or rare attributes may simply fall outside the axes we discover. In fact, perfectly capturing all underlying factors in a complex, real-world dataset is generally intractable. Nevertheless, our method still identifies diverse, meaningful concepts, and extending coverage to additional or more fine-grained factors remains an important direction for our future work. 
Moreover, our framework depends on the quality and scope of the pretrained vision language model (VLM), so it can only discover concepts the VLM recognizes. Fortunately, as VLMs are improving rapidly and our method is not restricted by a specific VLM, we can adopt stronger models as they become available.

\subsection{Broader Impact}
Our approach can extract diverse visual concepts from images and reuse them to synthesize new content, which could pose privacy issues such as deepfake generation or unauthorized duplication of digital content.

\subsection{Additional Implementation Details}
\label{subsec:additional_imp_detail}
Table~\ref{tab:hyperparams_of_models} summarizes hyper-parameters for model architectures and training used in our experiments. For baselines, we follow the default hyper-parameters recommended by the official codes. All baselines used DDIM inversion with guidance of 7.5 and 50 inference steps. 

\begin{table}[h!]
\vspace{0.1in}
\caption{Hyperparameters used in our experiments.}
\vspace{0.1in}
\label{tab:hyperparams_of_models}
\scriptsize
\centering
\begin{tabular}{@{}llcccc@{}}
\toprule
General     &   Batch Size       & 32 \\
            &   Training Steps   & 100k \\
            &   Learning Rate    & 0.00003 \\    \midrule
Concept Encoder         &  Layers &     4    \\
                        &  Hidden Dim &     768    \\
                        &  Number of Heads  &     8    \\ \midrule
Regression Network          &  Layers &     768    \\
                        &  Input Dimension   &     768    \\
                        &  Hidden Dimensio   &     768    \\
                        &  Activation Function  &     ReLU \\ \bottomrule
\end{tabular}
\end{table}

\clearpage
\subsection{Prompt for Concept Axes Extraction}
\label{subsec:prompt_analysis}

\begin{figure*}[ht]
    \vspace{0.2in}
    \centering
    \includegraphics[width=1.0\linewidth]{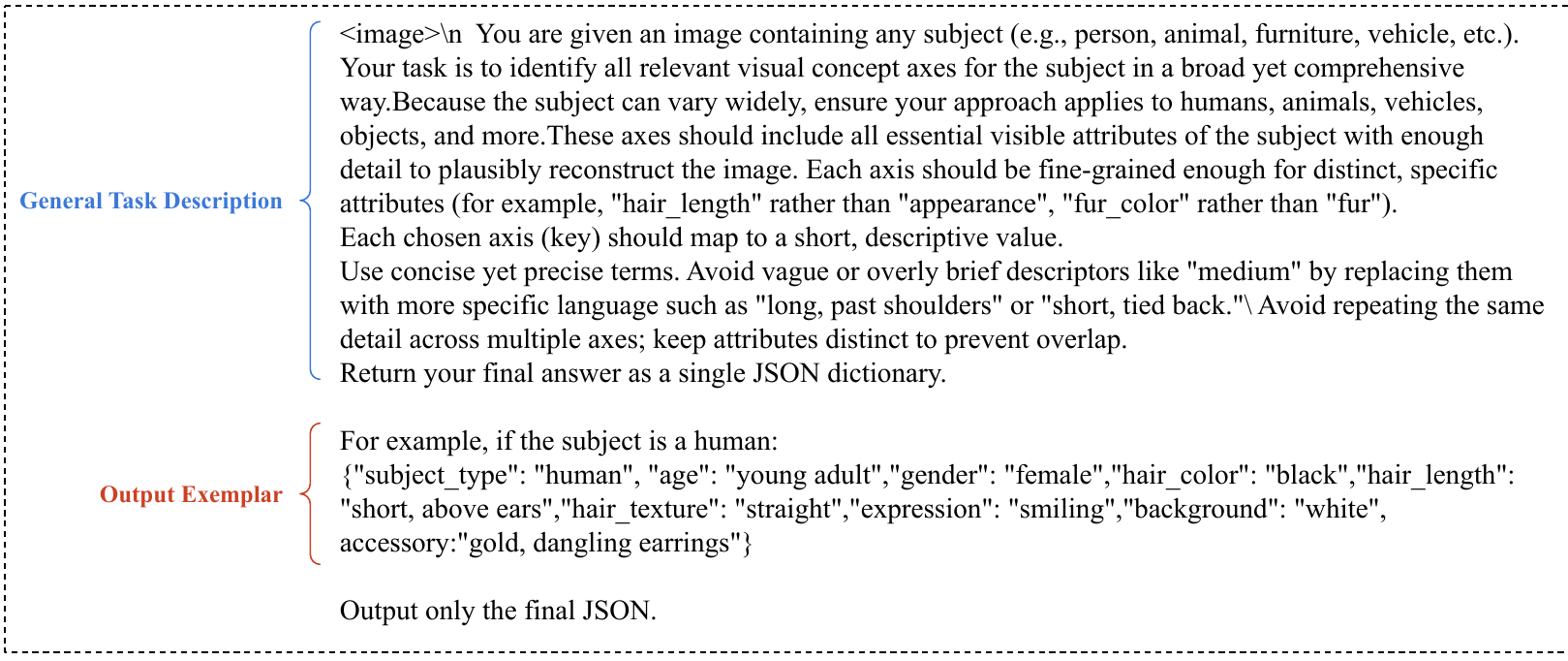}
    \vspace{-0.1in}
    \caption{Our complete prompt consists of a general task description and output exemplar.}
    \label{fig:complete_prompt}
\end{figure*}

\begin{figure*}[ht]
    \vspace{0.3in}
    \centering
    \includegraphics[width=1.0\linewidth]{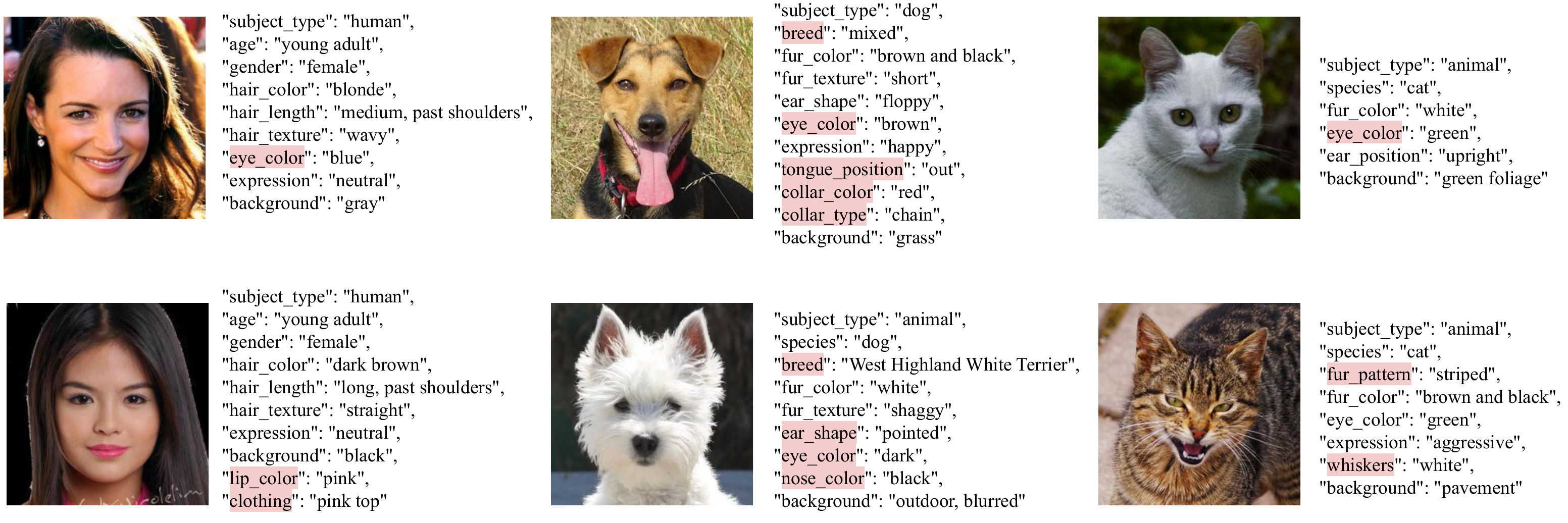}
    \caption{Examples of outputs from the VLM. Concept axes colored in red are unseen from the given exemplar.  
}
    \label{fig:prompt_output}
    \vspace{0.25in}
\end{figure*}

We provide the complete prompt and examples of the discovered concept axes per image in Figure~\ref{fig:complete_prompt} and Figure~\ref{fig:prompt_output}, respectively. 
As shown in Figure~\ref{fig:prompt_output}, our prompt successfully steers the VLM to identify diverse concept axes across different datasets, even when using only a single output exemplar of a human face. 

\clearpage
\newpage
\subsection{Human Evaluation}
\label{subsec:human_eval}

\begin{figure*}[ht]
    \centering
    \includegraphics[width=0.8\linewidth]{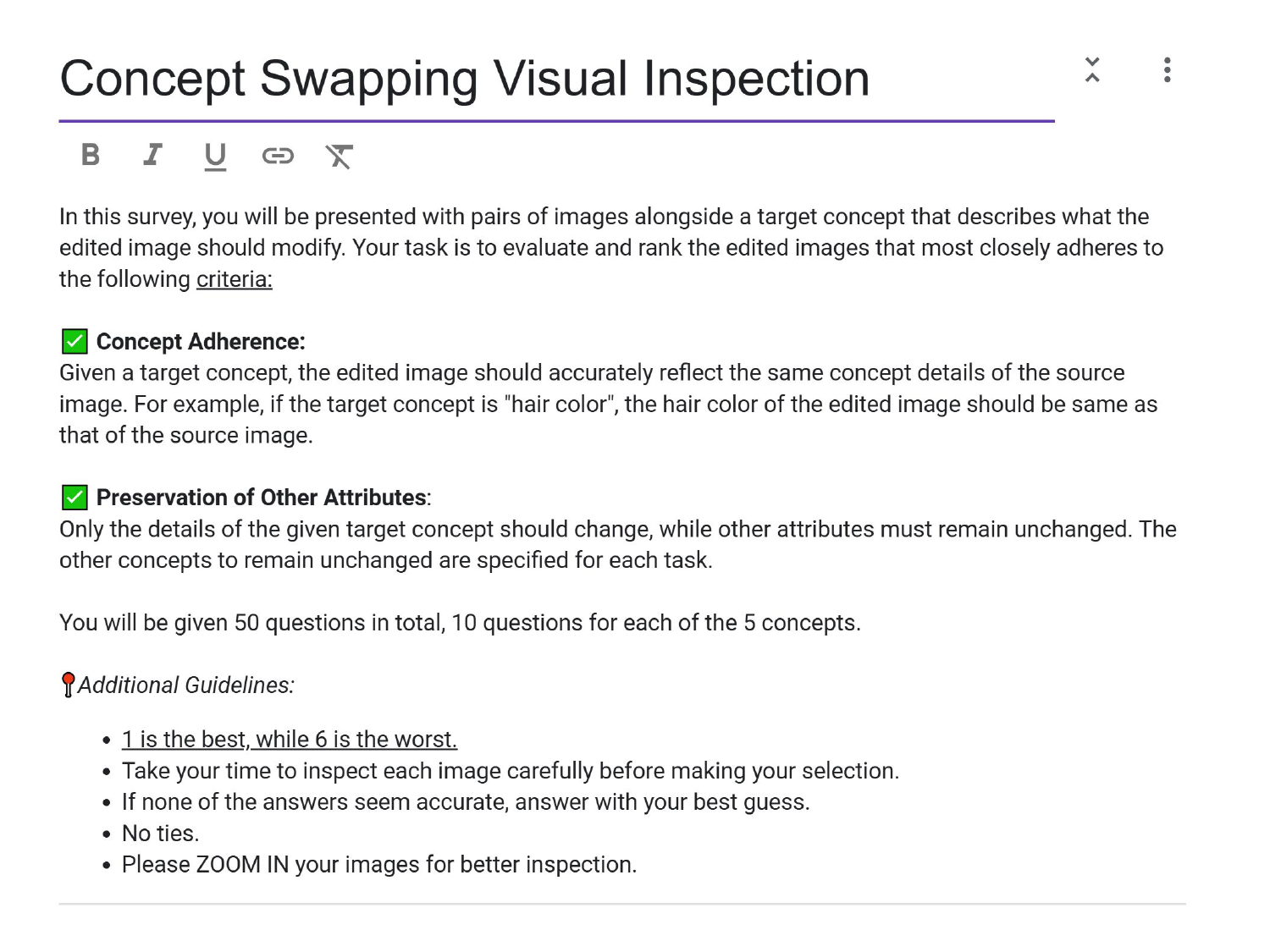}
    \vspace{-0.25in}
    \caption{General guidelines used in our human evaluation.
}
    \label{fig:human_guideline}
\end{figure*}

\begin{figure*}[ht]
    \centering
    \vspace{0.2in}
    \includegraphics[width=0.8\linewidth]{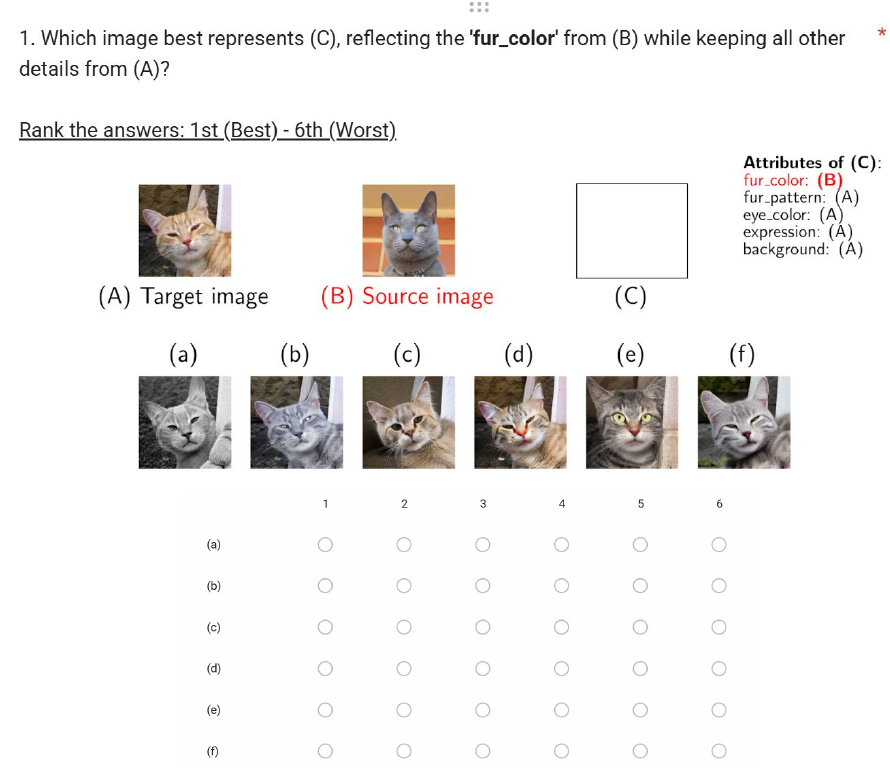}
\caption{Questionnaires used in human evaluation.
}
    \label{fig:human_question}
\end{figure*}

For human evaluation, we randomly select 10 pairs of images for each attribute. Then, we replace an attribute of one image with another one in each pair using each of the methods. 
We ensure that randomly selected attributes in each pair are different from each other so that the edited image is always recognizable. 
We collect 10 participants for each dataset (a total of 30) on Prolific~\cite{prolific} and provide a general guideline as in Figure~\ref{fig:human_guideline} for the task. Our questionnaire (Figure~\ref{fig:human_question}) asks participants to rank the images that most closely adhere to the criteria provided in our guideline. Following \cite{lee2023livcl}, we used Borda score metrics~\cite{saari2023selecting} to differentiate the scores according to each ranking, and final scores are normalized to a 0-1 scale.

\clearpage

\newpage
\subsection{Additional Qualitative Results}
\label{subsec:more_qualitative}

\subsubsection{Additional Qualitative Comparisons on Visual Concept Editing}
\begin{figure*}[ht]
    \centering
    \includegraphics[width=1.0\linewidth]{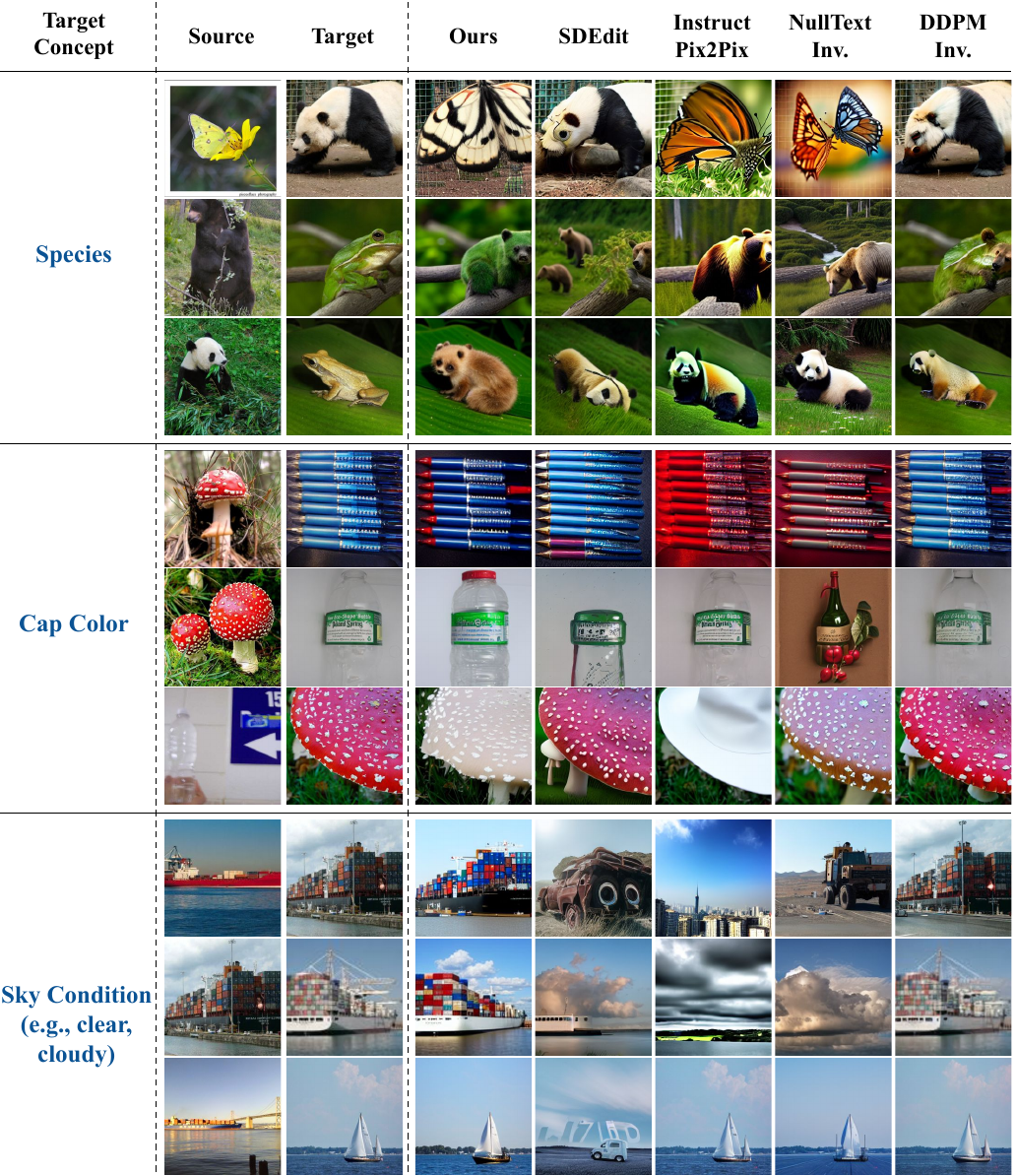}
    \caption{Additional qualitative comparison to baselines in ImageNet-S20
}
    \label{fig:more_imagenet_1}
\end{figure*}

\begin{figure*}[ht]
    \centering
    \includegraphics[width=1.0\linewidth]{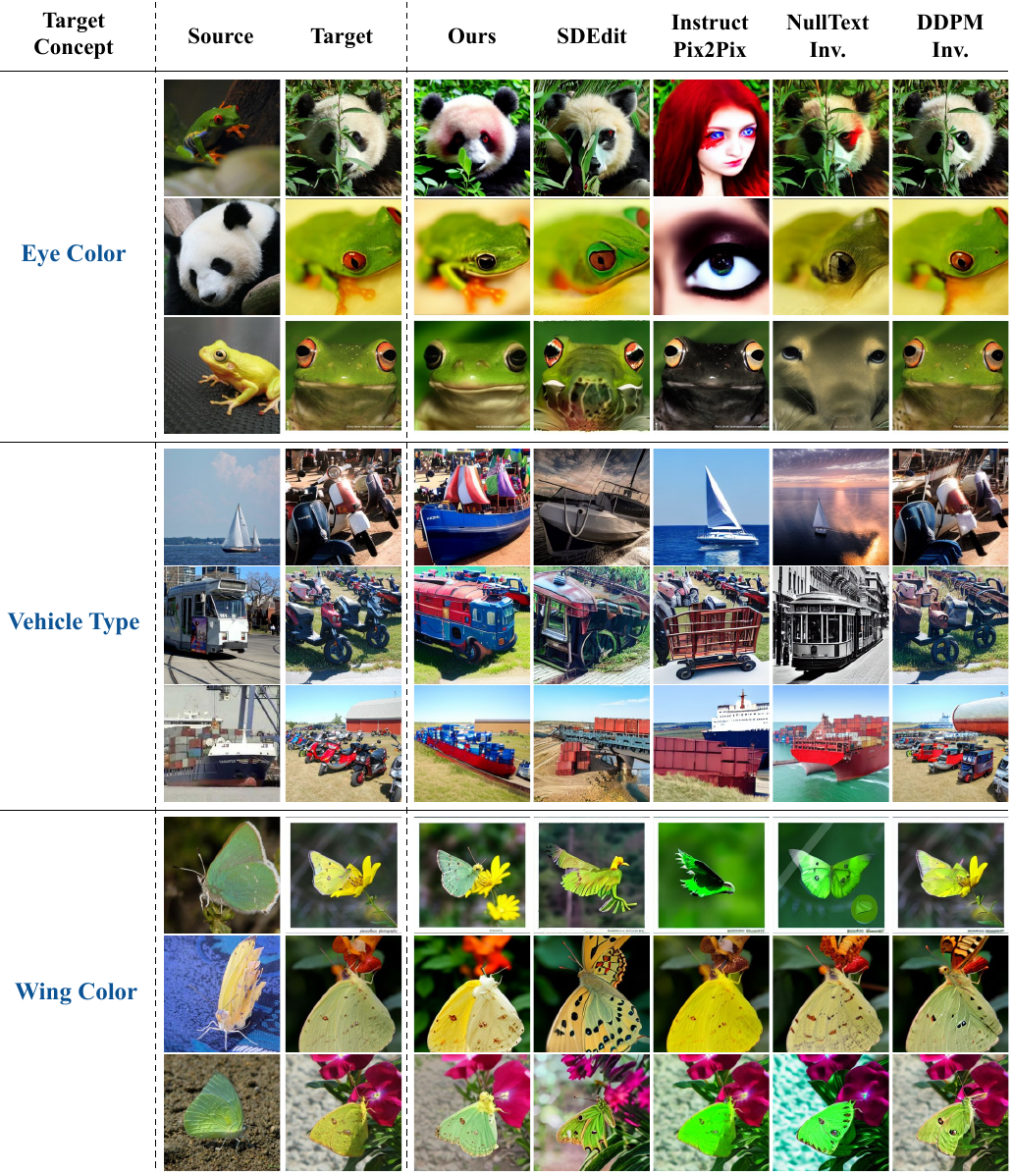}
    \caption{Additional qualitative comparison to baselines in ImageNet-S20
}
    \label{fig:more_imagenet_2}
\end{figure*}

\begin{figure*}[ht]
    \centering
    \includegraphics[width=1.0\linewidth]{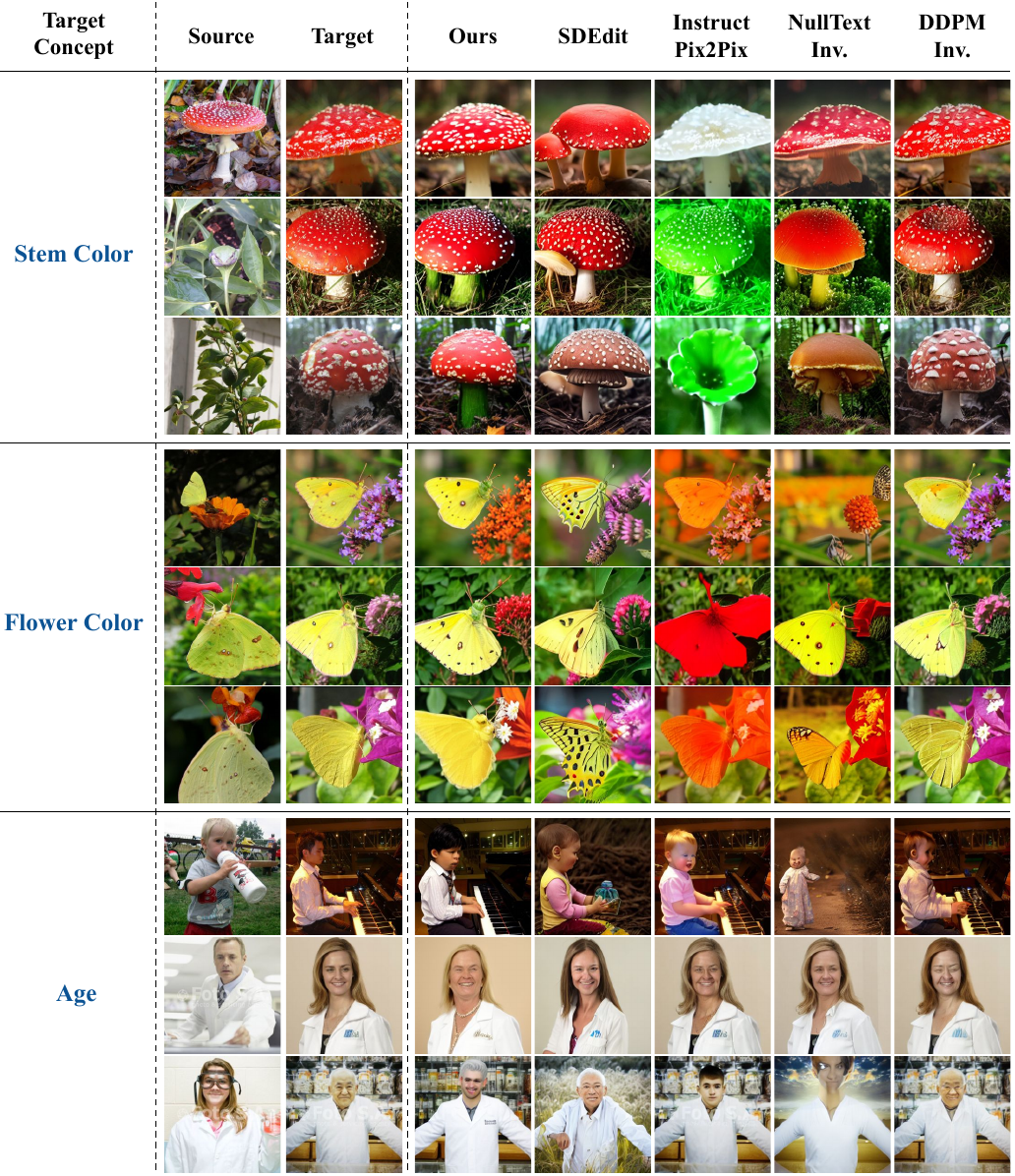}
    \caption{Additional qualitative comparison to baselines in ImageNet-S20
}
    \label{fig:more_imagenet_3}
\end{figure*}

\begin{figure*}[ht]
    \centering
    \includegraphics[width=1.0\linewidth]{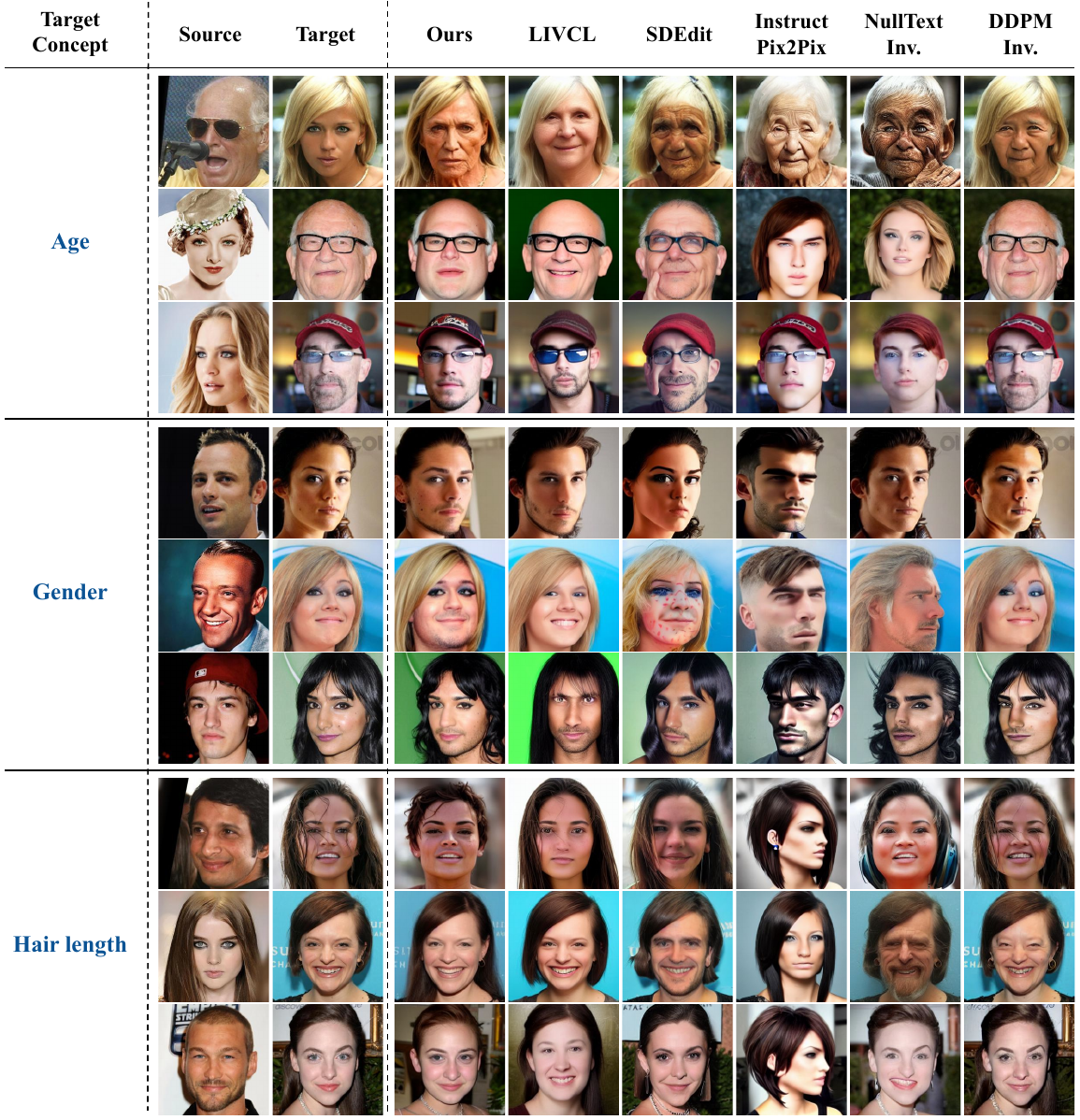}
    \caption{Additional qualitative comparison to baselines in CelebA-HQ
}
    \label{fig:more_qual_1}
\end{figure*}

\begin{figure*}[ht]
    \centering
    \includegraphics[width=1.0\linewidth]{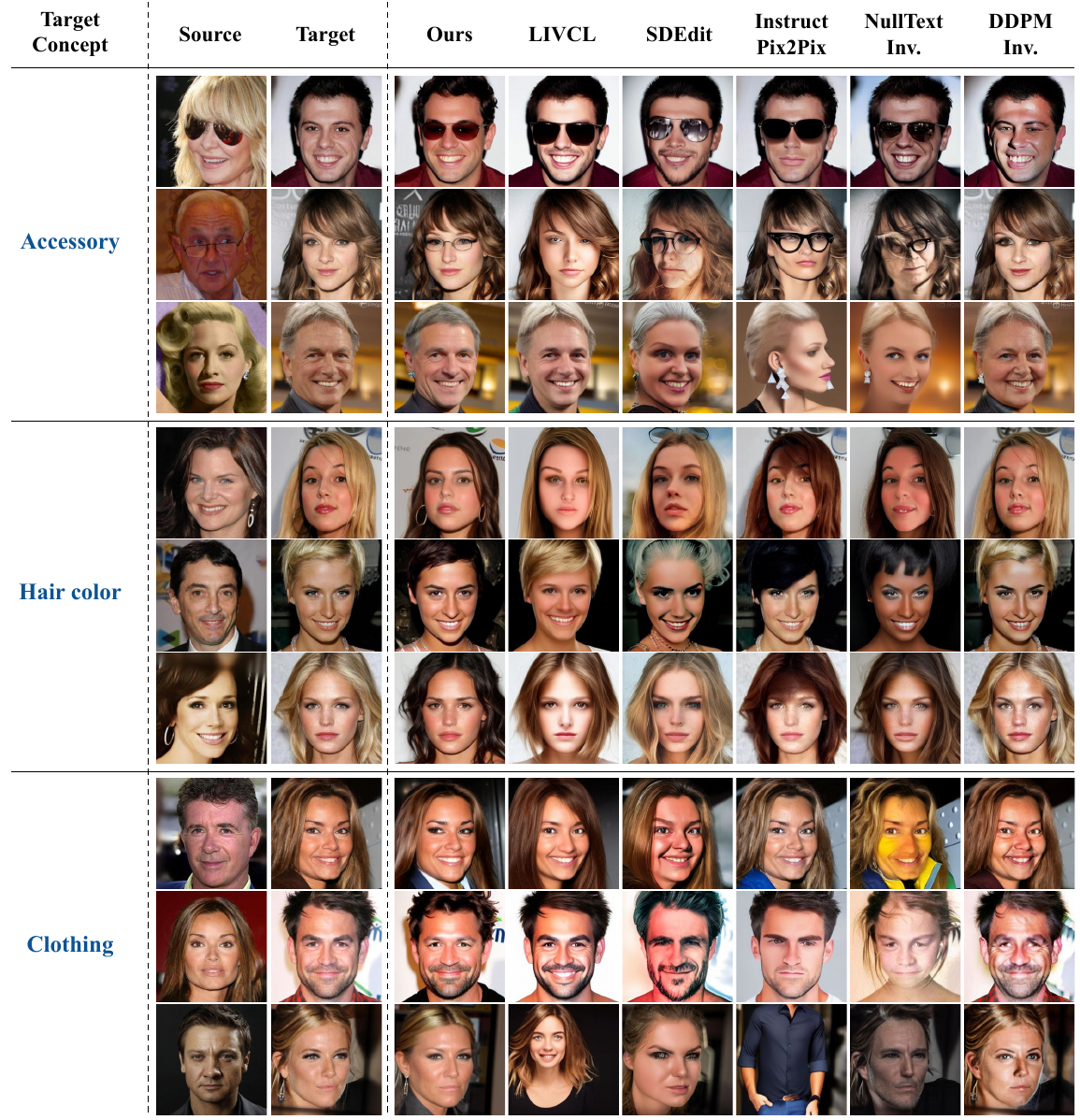}
    \caption{Additional qualitative comparison to baselines in CelebA-HQ
}
    \label{fig:more_qual_2}
\end{figure*}

\begin{figure*}[ht]
    \centering
    \includegraphics[width=1.0\linewidth]{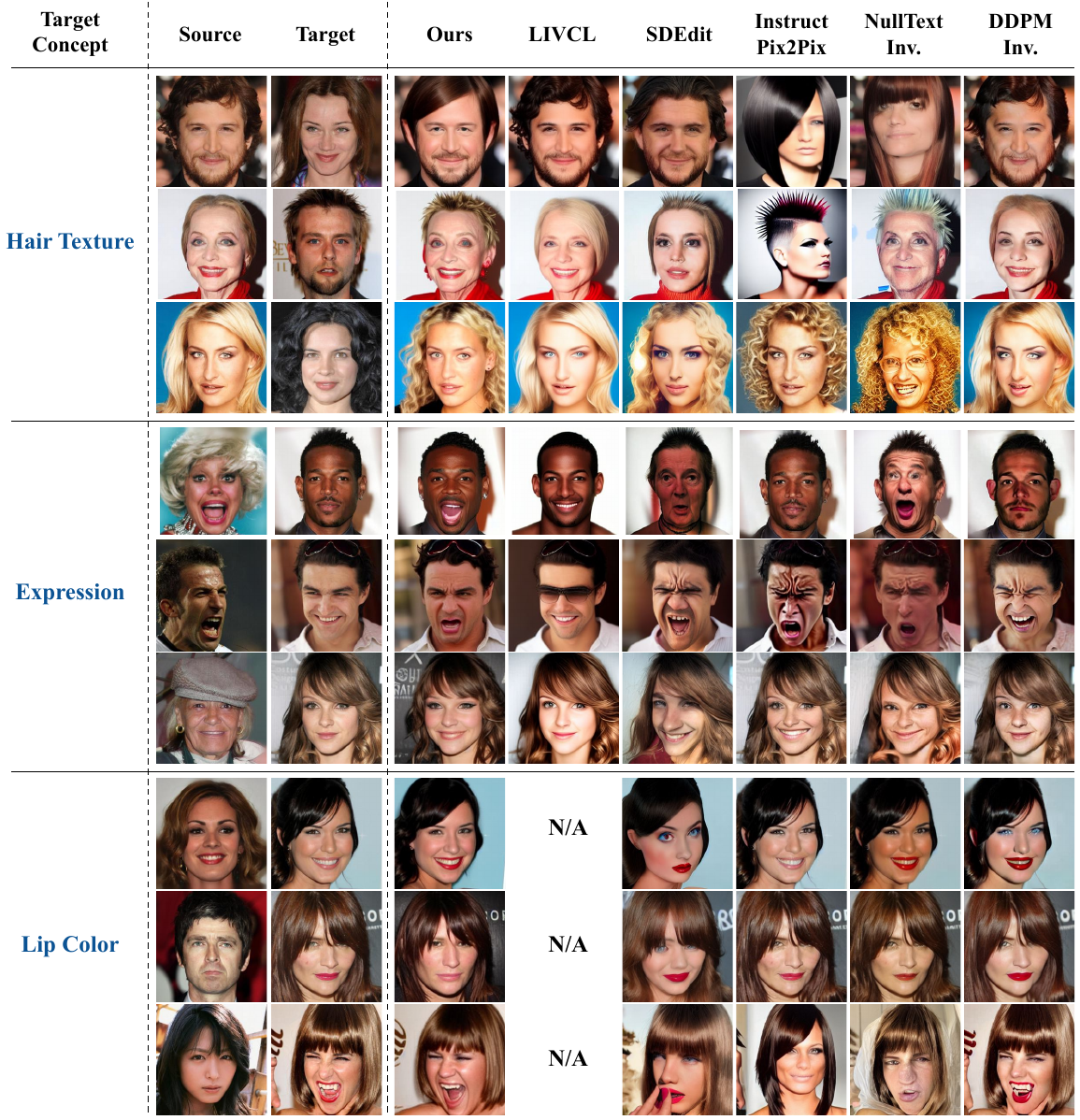}
    \caption{Additional qualitative comparison to baselines in CelebA-HQ
}
    \label{fig:more_qual_3}
\end{figure*}

\begin{figure*}[ht]
    \centering
    \includegraphics[width=1.0\linewidth]{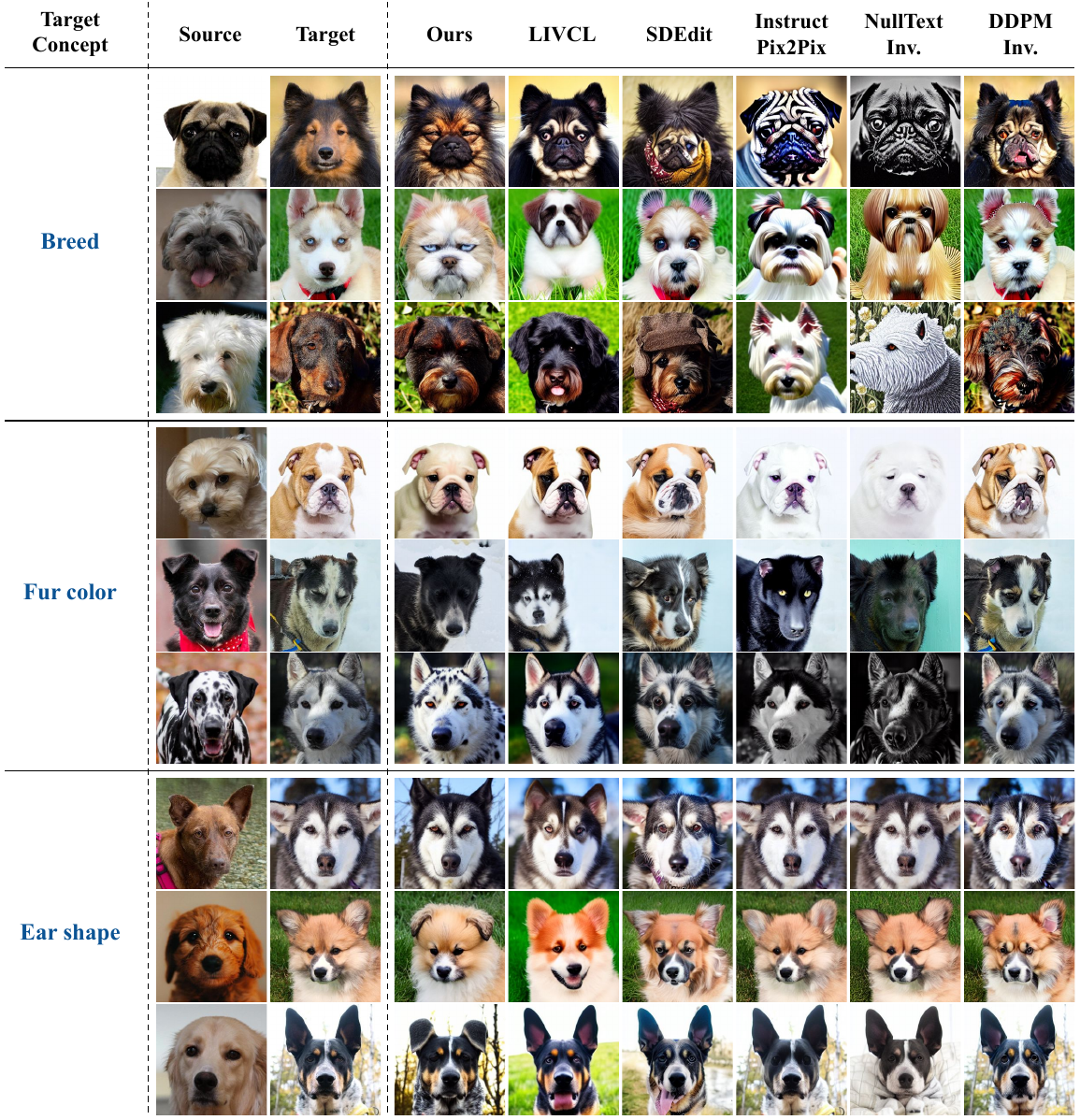}
    \caption{Additional qualitative comparison to baselines in AFHQ-Dog
}
    \label{fig:more_qual_4}
\end{figure*}

\begin{figure*}[ht]
    \centering
    \includegraphics[width=1.0\linewidth]{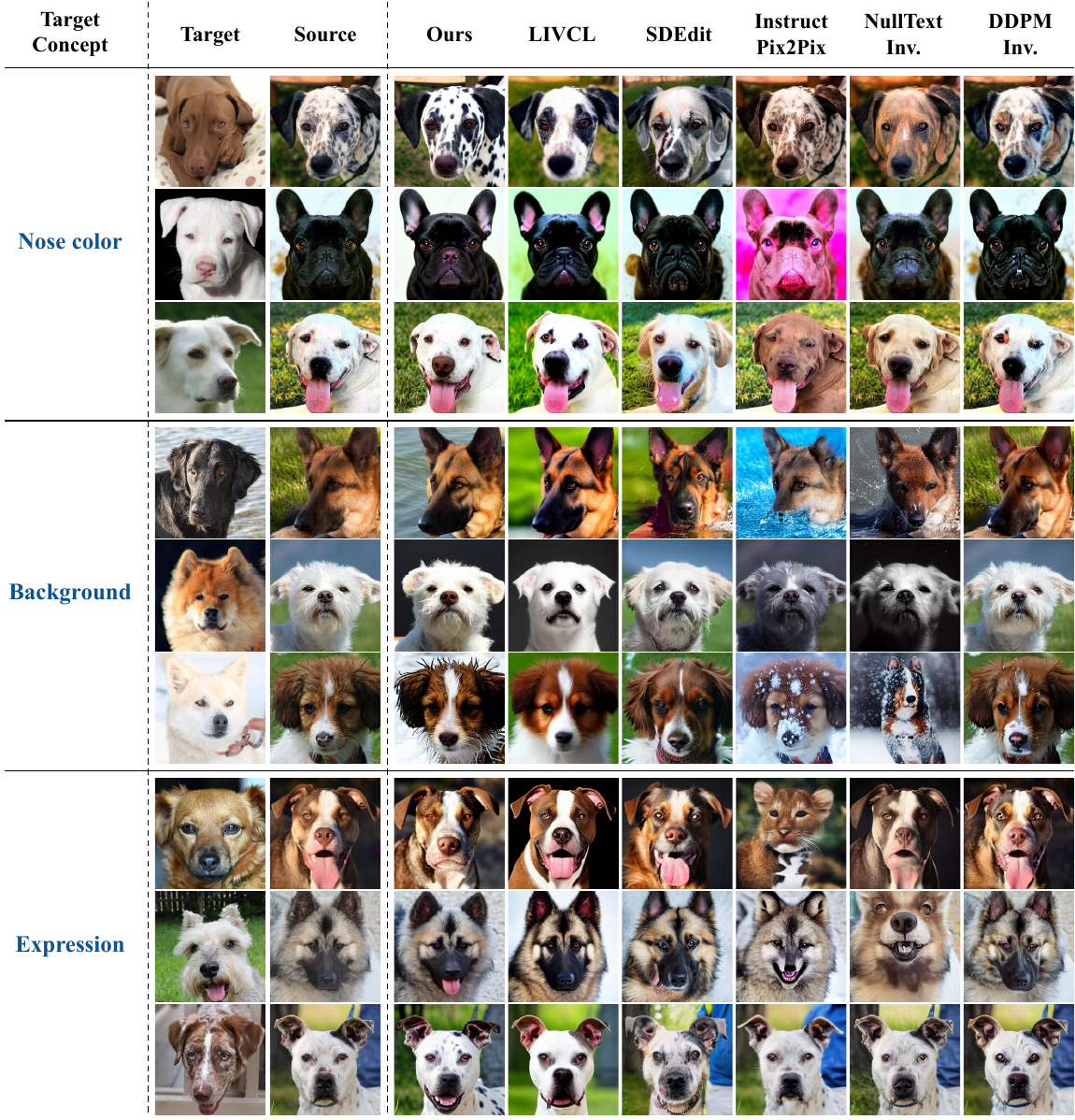}
    \caption{Additional qualitative comparison to baselines in AFHQ-Dog
}
    \label{fig:more_qual_5}
\end{figure*}

\begin{figure*}[ht]
    \centering
    \includegraphics[width=1.0\linewidth]{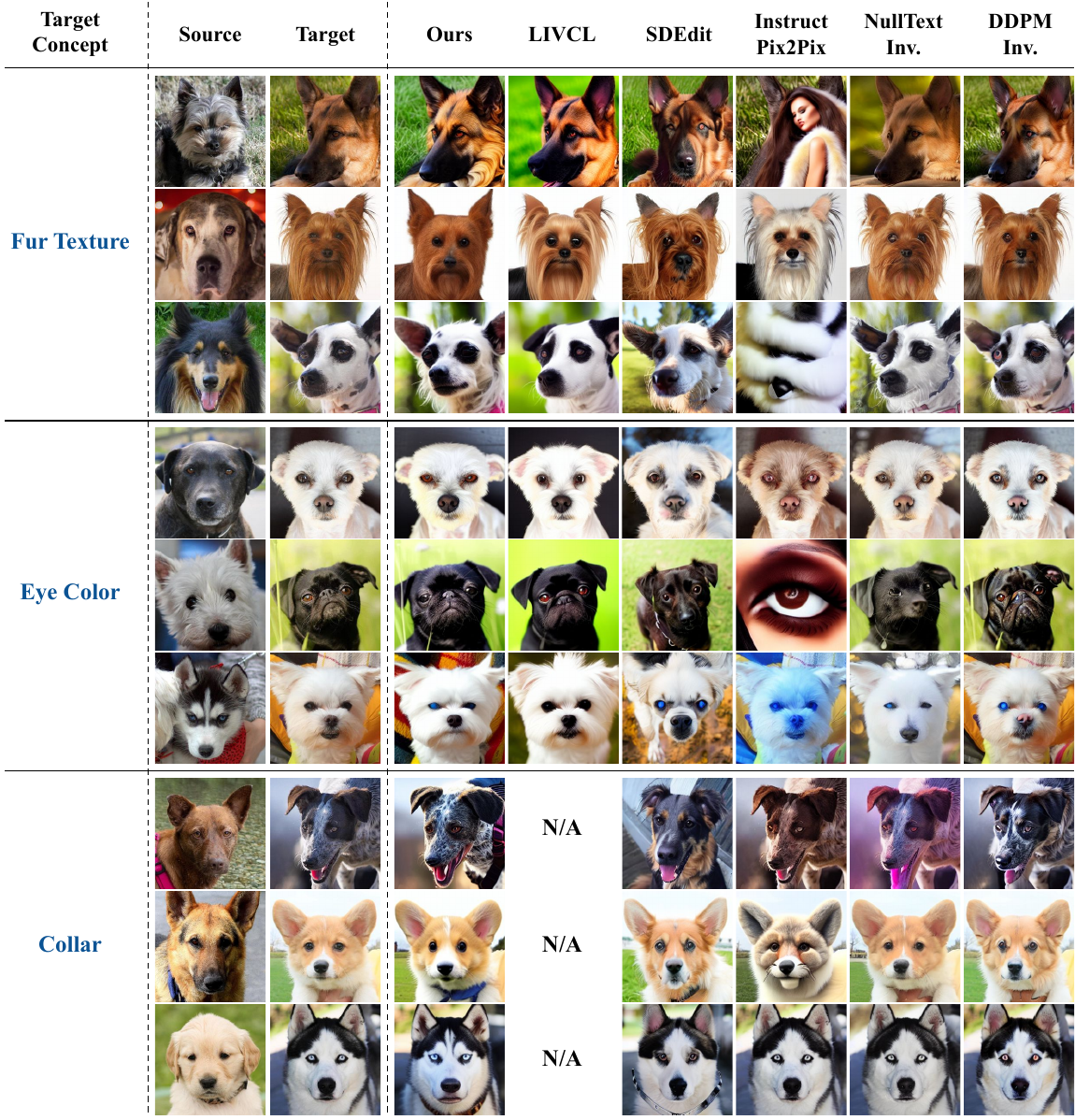}
    \caption{Additional qualitative comparison to baselines in AFHQ-Dog
}
    \label{fig:more_qual_6}
\end{figure*}

\begin{figure*}[ht]
    \centering
    \includegraphics[width=1.0\linewidth]{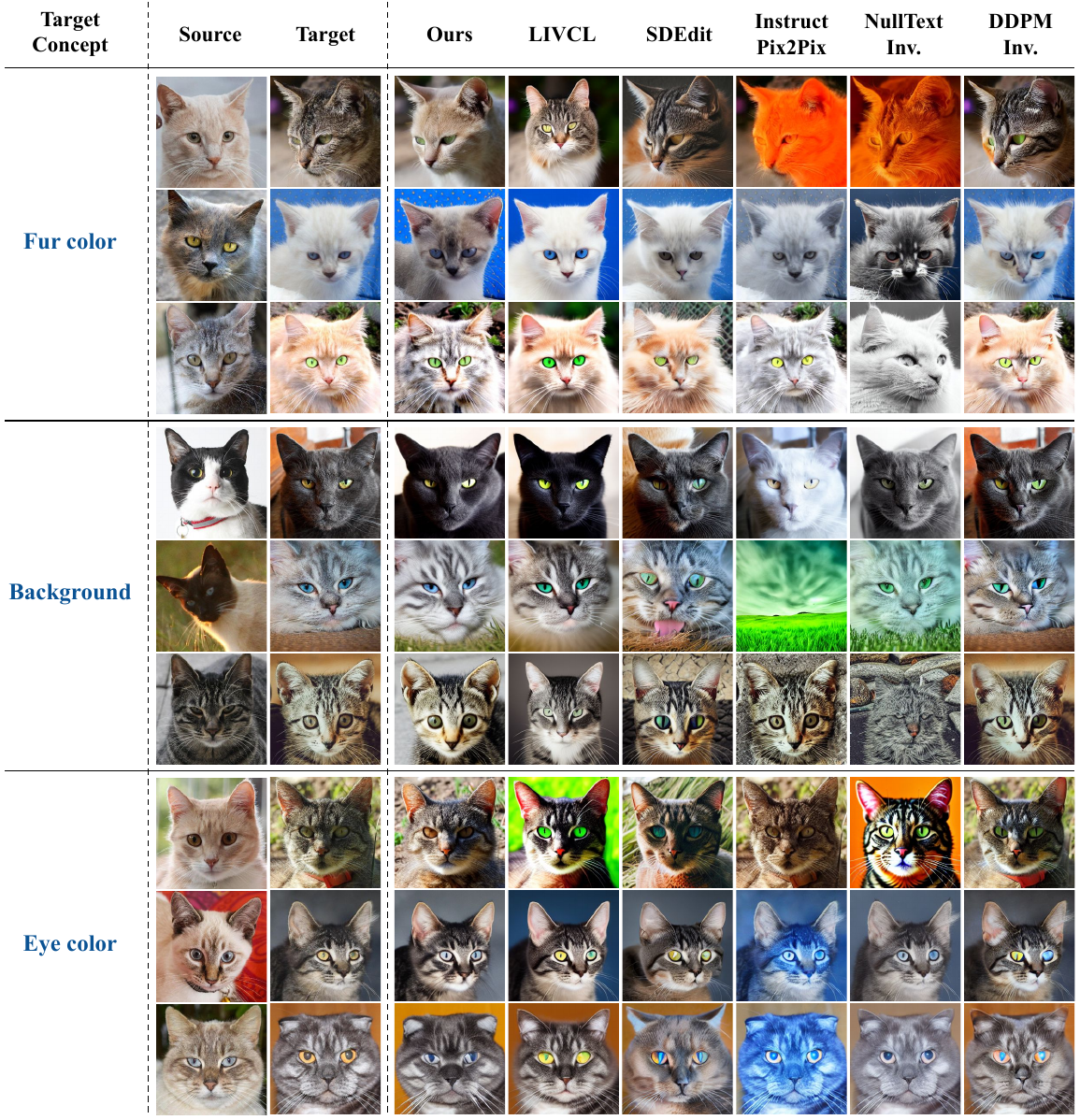}
    \caption{Additional qualitative comparison to baselines in AFHQ-Cat
}
    \label{fig:more_qual_7}
\end{figure*}

Figures~\ref{fig:more_imagenet_1}--\ref{fig:more_qual_7} present additional qualitative results along diverse concept axes discovered in ImageNet-S20, CelebA-HQ, and AFHQ datasets. Across all axes, our method consistently outperforms the baselines. Whereas the baselines often fail to accurately capture and transfer the specified visual attributes, our approach reliably extracts the visual concept from the source and transfers it to the target image. Since LIVCL trains a set of separate encoders only for the top–10 frequent axes, it was unable to evaluate “lip color” in Figure~\ref{fig:more_qual_3} and “collar” in Figure~\ref{fig:more_qual_6}, which are not among the top–10 most frequent concepts in the dataset, and we therefore mark those entries as N/A.

\vspace{0.2in}

\begin{figure*}[ht]
    \centering
    \includegraphics[width=1.0\linewidth]{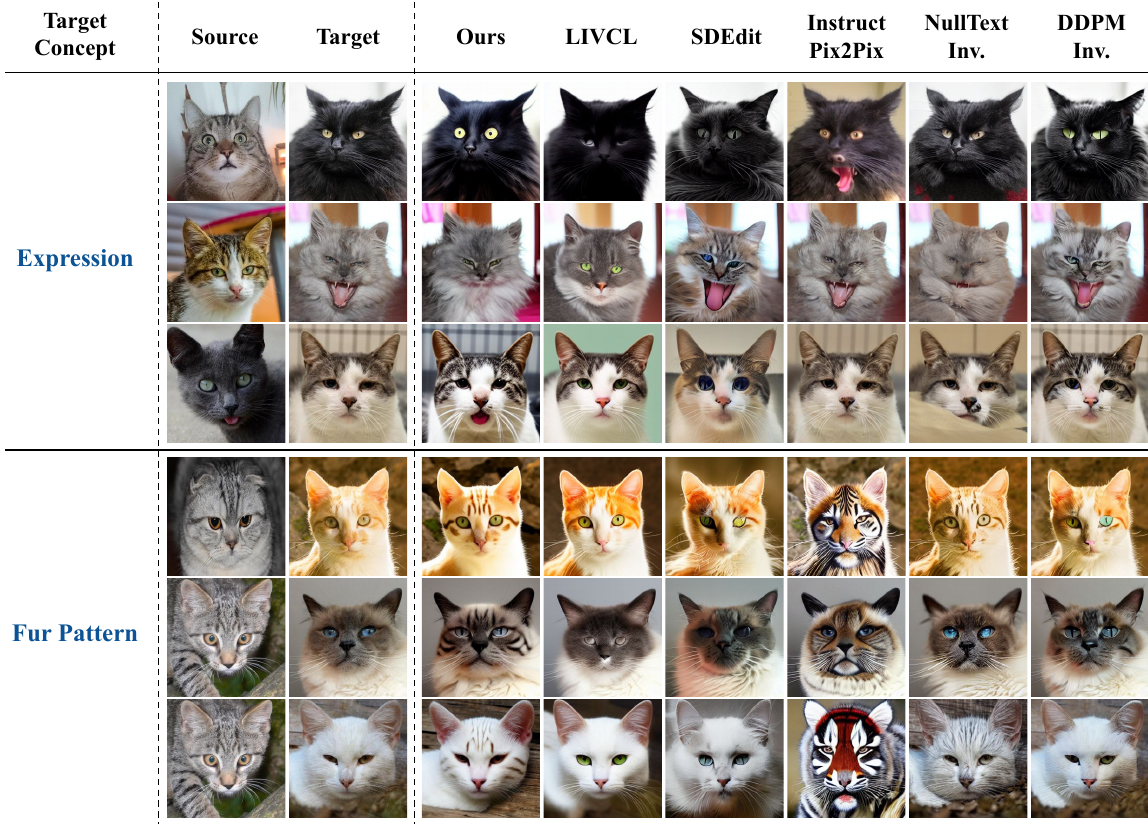}    
    \caption{Additional qualitative comparison to baselines in AFHQ-Cat
}
    \label{fig:more_qual_8}
\end{figure*}

\clearpage
\subsubsection{More Qualitative Results on Compositions from Multiple Images}
We provide more qualitative results on the composition of visual concepts from multiple images in Figure~\ref{fig:nuance_appendix1}-\ref{fig:nuance_appendix5}. 
We extract $N$ distinct visual concepts from $N$ different images and replace the corresponding visual concepts of the target images with them. Our method successfully transfers multiple visual concepts to target images, which implies that each visual concept extracted from source images is disentangled along other axes. 

\vspace{0.25in}
\begin{figure*}[ht]
    \centering
    \includegraphics[width=1.0\linewidth]{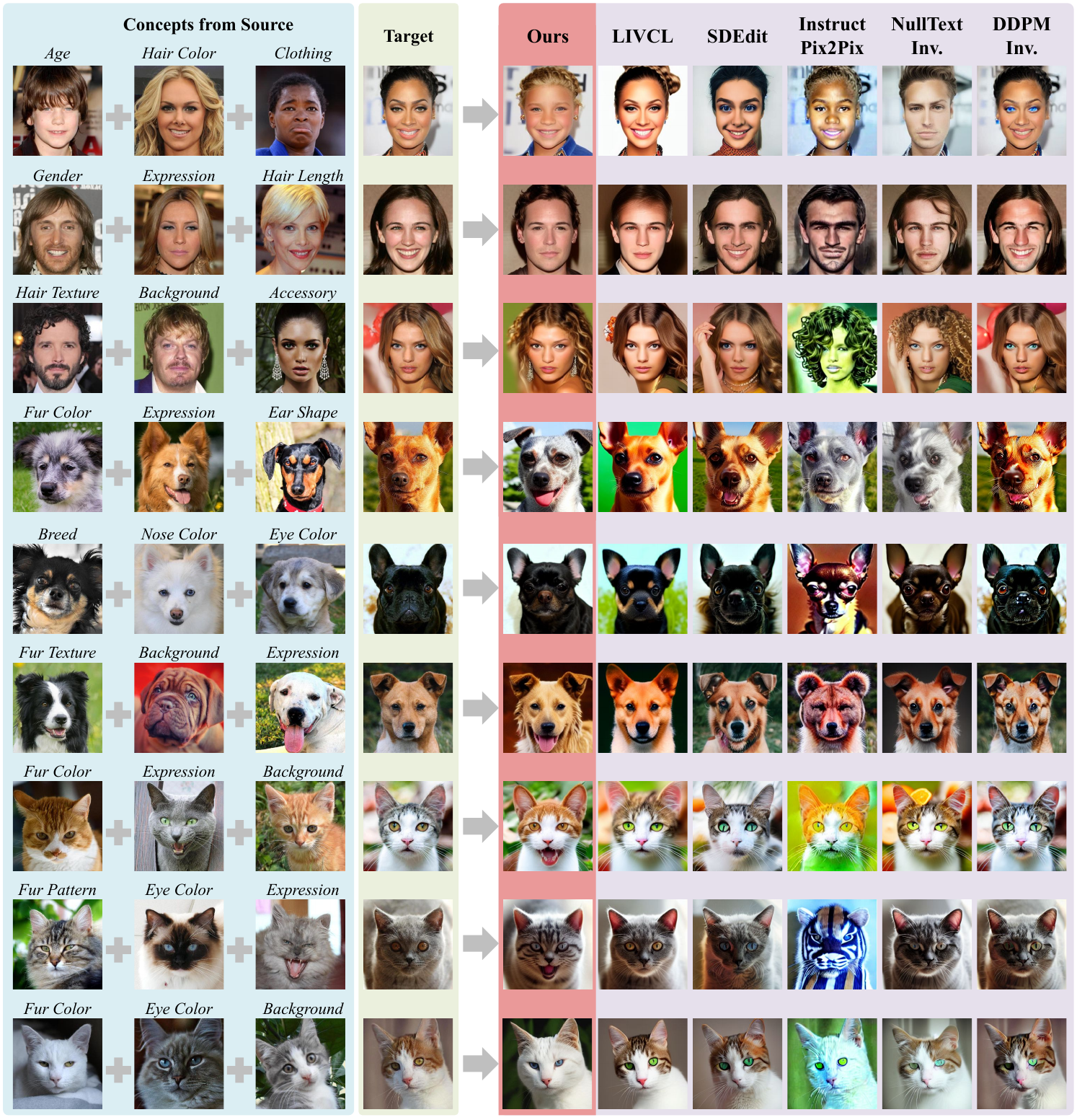}
    \caption{Compositions of visual concepts from multiple images ($N=3$).
}
    \label{fig:mult_comp_n=3}
\end{figure*}
\vspace{0.25in}

\begin{figure*}[ht]
    \centering
    \includegraphics[width=1.0\linewidth]{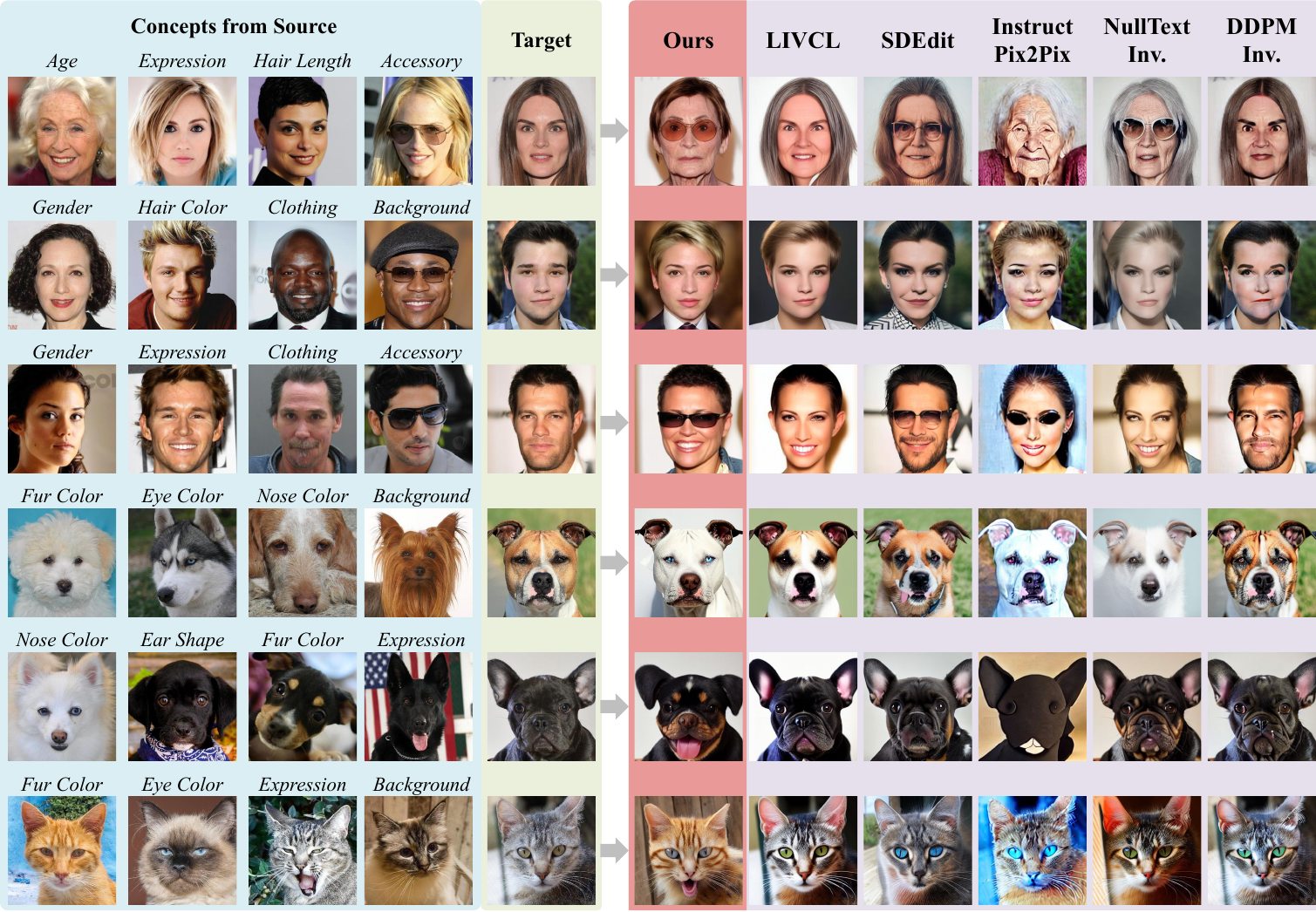}
    \caption{Compositions of visual concepts from multiple images ($N=4$).
}
    \label{fig:mult_comp_n=4}
\end{figure*}
\vspace{0.25in}

\clearpage
\subsubsection{More Qualitative Results on Visual Nuance Transfer}
\begin{figure*}[ht]
    \centering
    \includegraphics[width=1.0\linewidth]{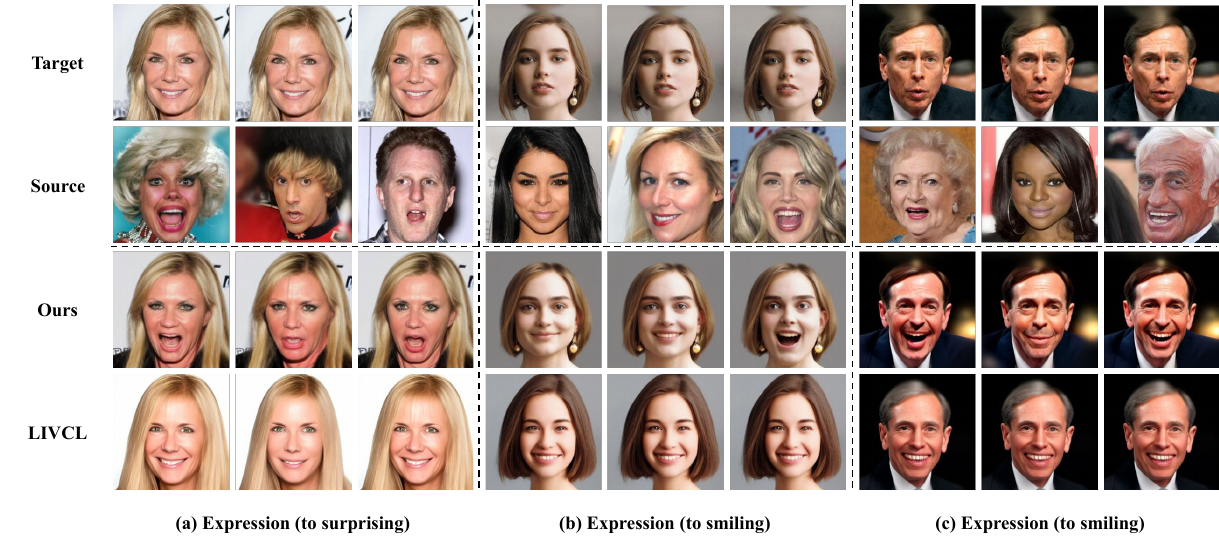}
    \caption{Transferring Visual Nuances from source to target images
}
    \label{fig:nuance_appendix1}
\end{figure*}

\begin{figure*}[ht]
    \centering
    \includegraphics[width=1.0\linewidth]{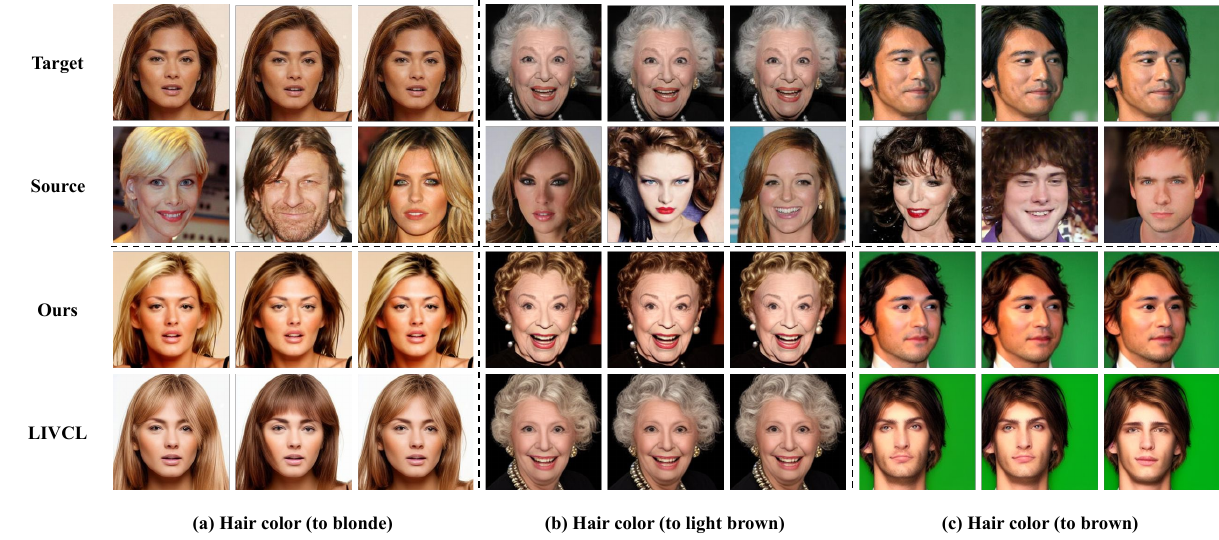}
    \caption{Transferring Visual Nuances from source to target images.  
}
    \label{fig:nuance_appendix2}
\end{figure*}

\begin{figure*}[ht]
    \centering
    \includegraphics[width=1.0\linewidth]{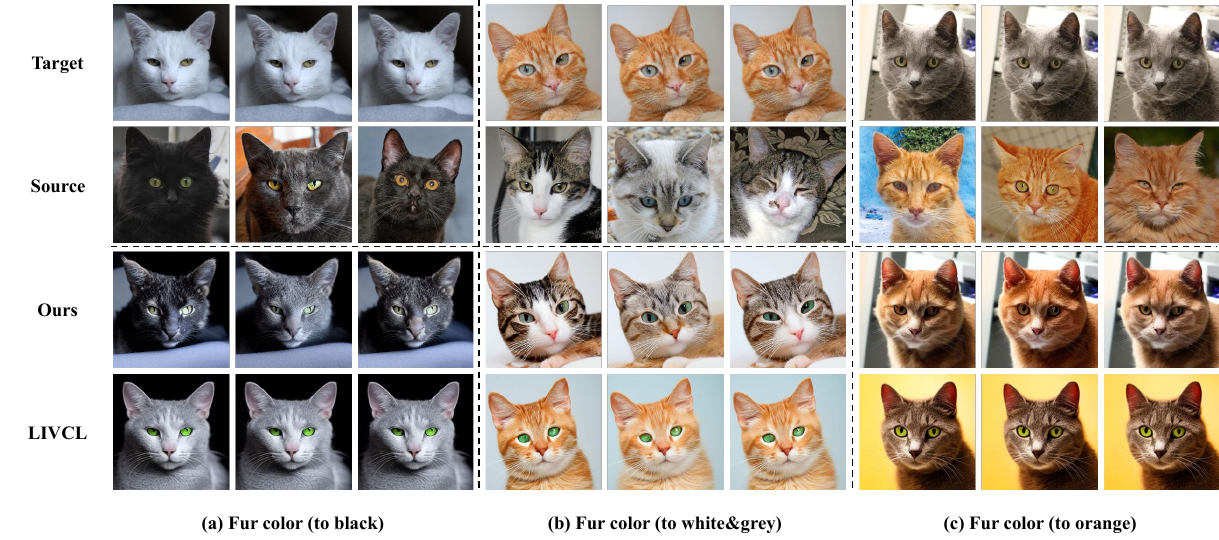}
    \caption{Transferring Visual Nuances from source to target images. 
}
    \label{fig:nuance_appendix3}
\end{figure*}

\begin{figure*}[ht]
    \centering
    \includegraphics[width=1.0\linewidth]{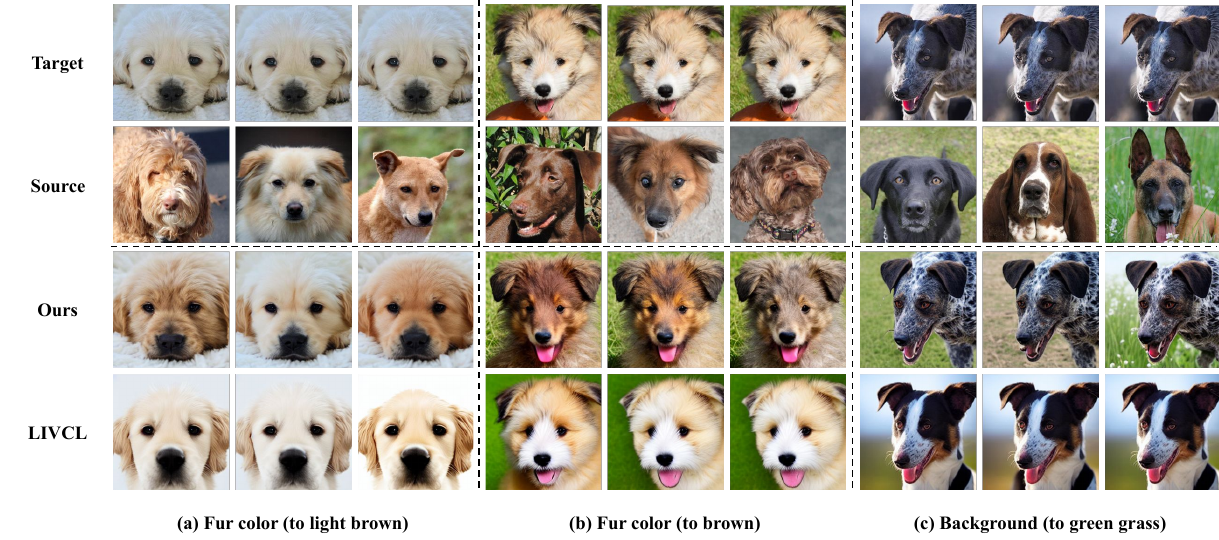}
    \caption{Transferring Visual Nuances from source to target images. 
}
    \label{fig:nuance_appendix4}
\end{figure*}

\begin{figure*}[ht]
    \centering
    \includegraphics[width=1.0\linewidth]{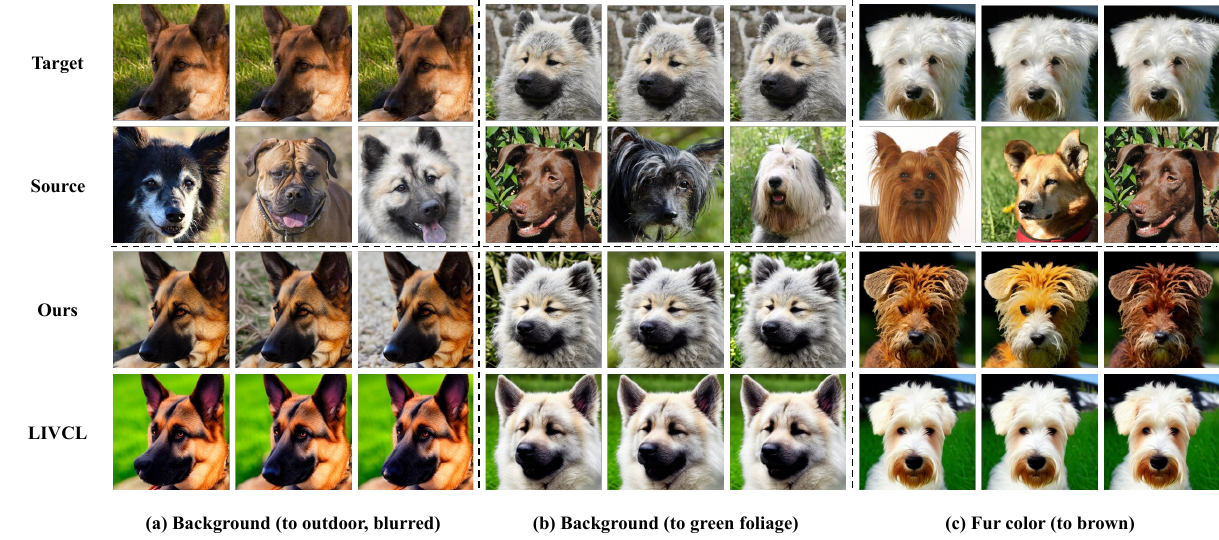}
    \caption{Transferring Visual Nuances from source to target images.
}
    \label{fig:nuance_appendix5}
\end{figure*}

We provide more qualitative results on transferring visual nuance from source to target images in Figure~\ref{fig:nuance_appendix1}-\ref{fig:nuance_appendix5}.

\vspace{0.25in}

\vspace{0.25in}

\vspace{0.25in}

\vspace{0.2in}

\vspace{0.2in}

\subsection{Computing Resources}
All of our experiments are conducted on a GPU Server that consists of an Intel Xeon Gold 6230 CPU, 256GB RAM, and 8 NVIDIA RTX 6000 GPUs (with 48GB VRAM). It takes about 48 GPU hours for each dataset.

\end{appendix}

\clearpage
\newpage

\section*{NeurIPS Paper Checklist}

The checklist is designed to encourage best practices for responsible machine learning research, addressing issues of reproducibility, transparency, research ethics, and societal impact. Do not remove the checklist: {\bf The papers not including the checklist will be desk rejected.} The checklist should follow the references and follow the (optional) supplemental material.  The checklist does NOT count towards the page
limit. 

Please read the checklist guidelines carefully for information on how to answer these questions. For each question in the checklist:
\begin{itemize}
    \item You should answer \answerYes{}, \answerNo{}, or \answerNA{}.
    \item \answerNA{} means either that the question is Not Applicable for that particular paper or the relevant information is Not Available.
    \item Please provide a short (1–2 sentence) justification right after your answer (even for NA). 
\end{itemize}

{\bf The checklist answers are an integral part of your paper submission.} They are visible to the reviewers, area chairs, senior area chairs, and ethics reviewers. You will be asked to also include it (after eventual revisions) with the final version of your paper, and its final version will be published with the paper.

The reviewers of your paper will be asked to use the checklist as one of the factors in their evaluation. While "\answerYes{}" is generally preferable to "\answerNo{}", it is perfectly acceptable to answer "\answerNo{}" provided a proper justification is given (e.g., "error bars are not reported because it would be too computationally expensive" or "we were unable to find the license for the dataset we used"). In general, answering "\answerNo{}" or "\answerNA{}" is not grounds for rejection. While the questions are phrased in a binary way, we acknowledge that the true answer is often more nuanced, so please just use your best judgment and write a justification to elaborate. All supporting evidence can appear either in the main paper or the supplemental material, provided in appendix. If you answer \answerYes{} to a question, in the justification please point to the section(s) where related material for the question can be found.

IMPORTANT, please:
\begin{itemize}
    \item {\bf Delete this instruction block, but keep the section heading ``NeurIPS Paper Checklist"},
    \item  {\bf Keep the checklist subsection headings, questions/answers and guidelines below.}
    \item {\bf Do not modify the questions and only use the provided macros for your answers}.
\end{itemize}


\begin{enumerate}

\item {\bf Claims}
    \item[] Question: Do the main claims made in the abstract and introduction accurately reflect the paper's contributions and scope?
    \item[] Answer: \answerYes{} 
    \item[] Justification: We states our motivation, contributions, scope of our work in abstract and introduction. 

\item {\bf Limitations}
    \item[] Question: Does the paper discuss the limitations of the work performed by the authors?
    \item[] Answer: \answerYes{} 
    \item[] Justification: We provide limitation of our work in Appendix. 

\item {\bf Theory assumptions and proofs}
    \item[] Question: For each theoretical result, does the paper provide the full set of assumptions and a complete (and correct) proof?
    \item[] Answer: \answerNA{} 
    \item[] Justification: We do not claim for theoretical results. 

    \item {\bf Experimental result reproducibility}
    \item[] Question: Does the paper fully disclose all the information needed to reproduce the main experimental results of the paper to the extent that it affects the main claims and/or conclusions of the paper (regardless of whether the code and data are provided or not)?
    \item[] Answer: \answerYes{} 
    \item[] Justification: We provide all the information needed to reproduce the experimental results. 

\item {\bf Open access to data and code}
    \item[] Question: Does the paper provide open access to the data and code, with sufficient instructions to faithfully reproduce the main experimental results, as described in supplemental material?
    \item[] Answer: \answerNo{}, 
    \item[] Justification: Our code is not cleaned and prepared enough for sharing. 

\item {\bf Experimental setting/details}
    \item[] Question: Does the paper specify all the training and test details (e.g., data splits, hyperparameters, how they were chosen, type of optimizer, etc.) necessary to understand the results?
    \item[] Answer: \answerYes{} 
    \item[] Justification: We specify all the details for experimental setting.

\item {\bf Experiment statistical significance}
    \item[] Question: Does the paper report error bars suitably and correctly defined or other appropriate information about the statistical significance of the experiments?
    \item[] Answer: \answerNo{}{} 
    \item[] Justification: Since our method requires costly GPU cost and time in training the diffusion model on real images, we were not affordable to conduct and provide repetitive experiments. We will add it in future. 

\item {\bf Experiments compute resources}
    \item[] Question: For each experiment, does the paper provide sufficient information on the computer resources (type of compute workers, memory, time of execution) needed to reproduce the experiments?
    \item[] Answer: \answerYes{}{} 
    \item[] Justification: We provide information of computing resources used for the experiments in Appendix. 
    
\item {\bf Code of ethics}
    \item[] Question: Does the research conducted in the paper conform, in every respect, with the NeurIPS Code of Ethics \url{https://neurips.cc/public/EthicsGuidelines}?
    \item[] Answer: \answerYes{}{} 
    \item[] Justification: We followed the NeurIPS Code of Ethics.

\item {\bf Broader impacts}
    \item[] Question: Does the paper discuss both potential positive societal impacts and negative societal impacts of the work performed?
    \item[] Answer: \answerYes{}{} 
    \item[] Justification: We discuss it in Appendix. 
    
\item {\bf Safeguards}
    \item[] Question: Does the paper describe safeguards that have been put in place for responsible release of data or models that have a high risk for misuse (e.g., pretrained language models, image generators, or scraped datasets)?
    \item[] Answer: \answerNo{} 
    \item[] Justification: : Our paper possess no risk

\item {\bf Licenses for existing assets}
    \item[] Question: Are the creators or original owners of assets (e.g., code, data, models), used in the paper, properly credited and are the license and terms of use explicitly mentioned and properly respected?
    \item[] Answer: \answerYes{} 
    \item[] Justification: We cite all the codes, data, paper, and pretrained model in our paper. 

\item {\bf New assets}
    \item[] Question: Are new assets introduced in the paper well documented and is the documentation provided alongside the assets?
    \item[] Answer: \answerNA{} 
    \item[] Justification: We do not release any new assets

\item {\bf Crowdsourcing and research with human subjects}
    \item[] Question: For crowdsourcing experiments and research with human subjects, does the paper include the full text of instructions given to participants and screenshots, if applicable, as well as details about compensation (if any)? 
    \item[] Answer: \answerYes{}{} 
    \item[] Justification: We provide detailed instructions of our human evaluation and we provide proper rewards to participants through Prolific website.

\item {\bf Institutional review board (IRB) approvals or equivalent for research with human subjects}
    \item[] Question: Does the paper describe potential risks incurred by study participants, whether such risks were disclosed to the subjects, and whether Institutional Review Board (IRB) approvals (or an equivalent approval/review based on the requirements of your country or institution) were obtained?
    \item[] Answer: \answerNA{}{} 
    \item[] Justification: Our method and human evaluation possess no risk. 

\item {\bf Declaration of LLM usage}
    \item[] Question: Does the paper describe the usage of LLMs if it is an important, original, or non-standard component of the core methods in this research? Note that if the LLM is used only for writing, editing, or formatting purposes and does not impact the core methodology, scientific rigorousness, or originality of the research, declaration is not required.
    \item[] Answer: \answerYes{}{} 
    \item[] Justification: We used VLM for automatic extraction of visual concepts and provide detailed information in the paper. 

\end{enumerate}
\end{document}